\documentclass[conference]{IEEEtran}
\usepackage{times}

\IEEEoverridecommandlockouts                              % This command is only
\usepackage{xcolor}
\usepackage{comment}
\usepackage{graphics} % for pdf, bitmapped graphics files
\usepackage{epsfig} % for postscript graphics files
\usepackage{amsmath} % assumes amsmath package installed
\usepackage{amssymb}  % assumes amsmath package installed
\usepackage{mathtools}
\usepackage{marginnote}
\usepackage{hyperref}
\usepackage{cite}
\usepackage{mathrsfs}
\usepackage{algorithm}
\usepackage{bm}
\usepackage{algpseudocode}
\usepackage{booktabs}
\usepackage{graphicx}
\usepackage{svg}
\usepackage{balance}
\title{
{\spaceskip=0.65\fontdimen2\font
PRIME: \ul{P}hysically-consistent \ul{R}obotic \ul{I}nertial and \ul{M}otion
\ul{E}stimation for Legged and Humanoid Robots}
}
\author{
Jiarong Kang$^{1}$, Kunzhao Ren$^{1}$, Tao Pang, and Xiaobin Xiong$^{1,2,\dagger}$\\
$^{1}$University of Wisconsin-Madison, 
$^{2}$Current affiliation: Shanghai Innovation Institute\\
$^{\dagger}$Corresponding author: xiaobin.xiong@sii.edu.cn \\
Project website: \href{https://jkangkjr.github.io/PRIME-project/}{https://jkangkjr.github.io/PRIME-project}
}
\begin{document}
\flushbottom
\newcommand{\KANG}[1]{{\color{blue} #1}}

\newcommand{\ul}[1]{\underline{#1}}

\newcommand{\block}[1]{\noindent{\textbf{#1}:}}
\newcommand{\emphhh}[1]{{\color{yellow} \textbf{#1}}}

\maketitle
\thispagestyle{empty}
\pagestyle{empty}

%%%%%%%%%%%%%%%%%%%%%%%%%%%%%%%%%%%%%%%%%%%%%%%%%%%%%%%%%%%%%%%%%%%%%%%%%%%%%%%%
\begin{abstract}
Humanoid and legged robots interact with the environment through intermittent contacts, making accurate motion estimation fundamentally dependent on reasoning about contact dynamics. However, standard sensing pipelines—whether based on onboard proprioception with Extended Kalman Filters (EKFs) or external motion capture systems—recover only kinematics, while contact forces, contact timing, and inertial parameters remain unobserved. As a result, purely kinematic reconstructions often violate rigid-body dynamics, particularly during contact-rich motions. To enable accurate motion estimation from onboard kinematics in real-world deployment, we propose PRIME (Physically-consistent Robotic Inertial and Motion Estimation), a Maximum A Posteriori (MAP) formulation that refines measured kinematics and actuator commands into a dynamically consistent trajectory while jointly estimating frictional contact forces and physically consistent inertial parameters. Our approach incorporates differentiable contact dynamics with smoothed complementarity constraints and an Anitescu-style friction model, yielding a smooth optimization problem that remains tractable across versatile contact transitions. We evaluate PRIME on contact-rich locomotion with quadrupedal robots and the Unitree G1 humanoid, demonstrating improved trajectory consistency and accurate inertial parameter identification. Beyond improving state estimation and feedback control with calibrated inertial parameters, PRIME produces force- and contact-annotated motion reconstructions from real robots in deployment, which can be used to provide high-quality data for downstream learning applications, including large-scale behavior modeling and robot foundation models.
\end{abstract}

\section{Introduction}

Humanoid and legged robots interact with their environment through intermittent contacts, where accurate motion estimation is fundamentally coupled with contact dynamics \cite{wensing2023optimization, carpentier2018multicontact}. During locomotion and manipulation, impacts, stance transitions, and frictional interactions induce forces that directly shape the motion of the robot and the environment therein. Reliable reasoning about such behaviors therefore requires estimates that are not only kinematically accurate but also dynamically consistent. This challenge becomes especially pronounced in contact-rich settings, which are central to real-world deployment of legged and humanoid systems.

Despite significant advances in sensing and estimation, most existing pipelines remain limited to kinematic reconstruction. Standard onboard proprioceptive estimators, typically based on Extended Kalman Filters (EKFs) \cite{EKF} and various smoothers \cite{yoon2023invariant, KangMHE_RAL}, as well as high-precision external motion capture systems \cite{ze2025twist2}, recover only the robot’s configuration and velocity, while contact forces, contact timing, and inertial parameters remain latent. As a result, reconstructed motions often violate rigid-body dynamics, particularly during phases involving intermittent contact events. These inconsistencies in the recovered robot state/data not only hinder downstream tasks such as planning and control \cite{yang2025impact, FastCIMPC, kurtz2026inverse}, but also prevents data-driven methods such as imitation learning \cite{chi2025diffusion, tsuji2025survey, sun2024prompt}, vision-language-action models \cite{kim2024openvla, ma2024survey} and robot foundation models \cite{li2024foundation, firoozi2025foundation} to function in real-world.

\begin{figure}[t]
    \centering
    \setlength{\abovecaptionskip}{0pt}
    \includegraphics[width=1.0\linewidth]{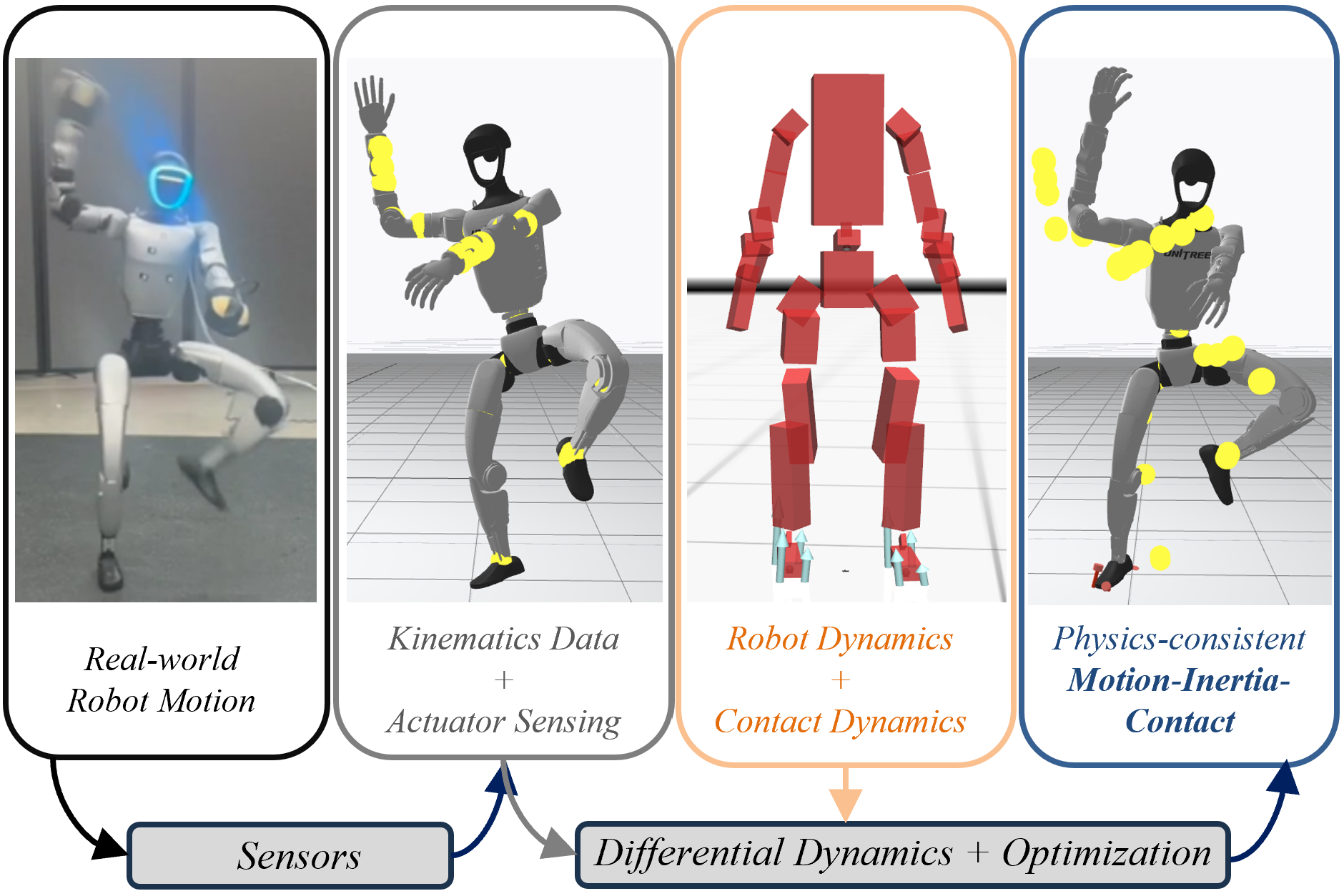}
    \vspace{-17pt}
    \caption{Overview of PRIME: PRIME reconstructs \textit{physically consistent robot trajectories} and \textit{hardware-matched inertial parameters} from kinematic measurements and onboard actuator sensing using a parameter full-information estimation framework with differentiable contact dynamics.}
    \label{Framework}
    \vspace{-23pt}
\end{figure}

To address this problem, we propose PRIME (Physically-consistent Robotic Inertial and Motion Estimation) as illustrated in Fig. \ref{Framework}, a framework for refining measured kinematics into dynamically consistent motion trajectories using IMU, joint sensors and motion capture measurements. PRIME formulates motion and parameter estimation as a Maximum A Posteriori (MAP) problem that jointly estimates state trajectories, frictional contact forces, and physically consistent inertial parameters. Rather than treating contacts as exogenous or implicitly absorbed into noise, PRIME explicitly reasons about contact dynamics, enabling physically grounded reconstruction of contact-rich motions. While our work here focuses on legged and humanoid locomotion, the formulation naturally extends to other contact-rich behaviors, such as robot manipulation, where PRIME could in principle estimate dynamically consistent robot-object motion, contact forces, and inertial parameters under intermittent and unobserved interactions.

PRIME incorporates differentiable contact dynamics with smoothed complementarity constraints \cite{TaoCIMPC} and an Anitescu-style friction model \cite{Anitescu}, yielding a smooth optimization problem that remains tractable across contact transitions. This formulation enables PRIME to handle switching contact modes robustly without requiring explicit contact sensing or discontinuous dynamics. The resulting problem is solved efficiently as a Parameter Full-Information Estimation \cite{IJRRLCP} task using Feasibility-Driven Differential Dynamic Programming (FDDP) \cite{FDDP}, refining noisy or dynamically inconsistent kinematics into trajectories that satisfy rigid-body dynamics and contact constraints, while simultaneously identifying inertial parameters consistent with the observed motion.

We evaluate PRIME on contact-rich locomotion across diverse legged platforms, including the Unitree G1 humanoid. The results show that PRIME produces dynamically consistent motion trajectories and accurate inertial-parameter estimates from kinematic sensing, while reconstructing high-quality contact interactions over long horizons (10~s to 20~s). These improvements enable more reliable state estimation and control, and provide physically grounded motion reconstructions for downstream analysis and learning methods that depend on high-fidelity data. Overall, the results suggest that explicitly reasoning about contact dynamics is a key enabler for accurate motion estimation in real-world robotic systems.

\section{Related Work}
% @article{IEKF,

State estimation and system identification for legged robots have been studied extensively, motivated by the challenges posed by intermittent ground contact, actuator uncertainties, and highly dynamic motions. For state estimation, classical approaches commonly rely on Kalman filters and their variants \cite{EKF} to fuse onboard sensors, including IMU measurements, joint encoders, and contact sensing, to achieve drift-reduced estimation. More recent optimization-based methods leverage factor-graph formulations or windowed optimization, such as fixed-lag smoothing~\cite{yoon2023invariant} and moving-horizon estimation~\cite{KangMHE_RAL}, and improve accuracy by exploiting a finite history of heterogeneous measurements from both proprioceptive sensors and exteroceptive modalities such as vision and Lidar \cite{FactorGraph}.

However, much of the existing work remains largely kinematics-centric. Contact is typically incorporated either through direct contact sensing~\cite{EKF} or through decoupled inference from actuator torque measurements. Such sensor-level cues usually provide only binary contact indicators, rather than contact-force estimates. Recent efforts have begun to incorporate dynamics-related factors into the estimation, for example by combining momentum terms with contact constraints to recover ground-reaction forces from onboard measurements \cite{ForceMHE}, or by integrating factor-graph estimation with a decoupled momentum observer to estimate torso disturbances~\cite{VIEW}. Nevertheless, contact is still often treated in a decoupled manner without enforcing full physical consistency, which can yield kinematic reconstructions that violate rigid-body dynamics and contact physics, thereby limiting the quality of data used in downstream data-driven pipelines.

Building on state estimation, system identification is typically formulated as a least-squares problem, with physical consistency enforced through constrained inertial parameterizations~\cite{wensing2024geometric}. However, when state estimation is largely restricted to kinematics-based reconstruction, contact interactions are not explicitly modeled and must therefore be addressed separately in the identification process. Existing approaches~\cite{linearID} either rely on high-resolution force/torque sensing at contact interfaces or perform identification within subspaces that reduce sensitivity to unknown external forces~\cite{NullID}. These approaches often assume relatively high-quality data and address noise primarily through pre-processing such as low-pass filtering, which can be insufficient under frequent contact transitions and unmodeled interaction forces.

Recent works have begun to combine state estimation and system identification in a unified framework, allowing both tasks to benefit from physical consistency. For example, \cite{PMHE_England} formulates a parameter moving-horizon estimation problem for legged robots and jointly estimates trajectories and inertial parameters that are consistent with a measured contact sequence, but the approach remains limited by decoupled contact measurements. To enforce dynamics-consistent contact, \cite{IJRRLCP} proposes a joint parameter and contact estimation method via batch optimization with explicit linear complementarity constraints, yielding physically consistent contact forces and inertial parameters. However, due to the numerical difficulties induced by complementarity constraints, the method is only validated in a single-contact scenario of planar systems. To mitigate these numerical issues, \cite{BoxpushDDP} introduces differentiable contact dynamics and demonstrates identification in box sliding experiments. More broadly, contact models that support derivative-based optimization have been widely used for locomotion and manipulation control \cite{TaoCIMPC, CIMPChand, MujocoMPC}, but remain less explored for joint estimation and identification, particularly in complex, distributed-contact settings such as humanoid locomotion. Another line of work improves contact fidelity by using sampling \cite{SampleMPC}, which performs identification by querying a simulator. However, due to the dimensionality limits of sampling, these methods typically sample only inertial parameters without refining the motion trajectory.

\section{Modeling}

These gaps in related works motivate our work PRIME. By leveraging smoothed contact dynamics with informative analytic gradients, PRIME robustly reconstructs physically-consistent trajectories, contact interactions, and inertial parameters over long horizons for high-dimensional legged systems. In this section, we present the modeling of dynamics and measurements that form the basis of PRIME. 
 
% PRIME aims to simultaneously estimate robot motion and uncertain inertia properties from available sensor data. In this section, we present the basic modeling of robot dynamics and sensor measurement. 

\subsection{Robot Dynamics Modeling}
The legged robot system is described by $\mathbf{x} = \begin{bmatrix} \mathbf{q}^\intercal & \mathbf{v}^\intercal \end{bmatrix}^\intercal$, where $\mathbf{q} \in SE(3) \times \mathbb{R}^n$ denotes the generalized coordinates, including the floating-base pose $\mathbf{p}$ and $\mathbf{R}$, and the joint positions $\boldsymbol{\alpha}$, with $n$ representing the number of articulated joints. The generalized velocity is given by $\mathbf{v} \in \mathfrak{se}(3) \times \mathbb{R}^n$, comprising both the base twist and joint velocities.  
The continuous-time dynamics is approximated by a discrete-time system using a semi-implicit Euler integration scheme:
\begin{equation}
\begin{aligned}
    \mathbf{q}^+ - \mathbf{q} &= \Delta t\, \mathbf{v}^+, \\
    \mathbf{M}(\mathbf{q})({\mathbf{v}}^{+} - {\mathbf{v}}) &= \Delta t \big(\mathbf{B} \mathbf{u} - \mathbf{h}(\mathbf{q}, \mathbf{v}) + \textstyle \sum_i \mathbf{J}_i(\mathbf{q})^\intercal \mathbf{f}_i \big),
\end{aligned}    
\label{eq:discrete_dynamics}
\end{equation}
where the term $\mathbf{M}(\mathbf{q}) \in \mathbb{R}^{(6+n)\times(6+n)}$ represent the inertia matrix, and $\mathbf{h}(\mathbf{q},\mathbf{v})$ is the bias vector including Coriolis, centrifugal and gravitational effects. $\Delta t$ is the time discretization, and the superscript $\square^+$ denotes the states at the next time step. The input matrix $\mathbf{B} \in \mathbb{R}^{(6+n)\times n}$ maps the joint torque vector $\mathbf{u}$ to the generalized force space. $\mathbf{f}_i \in \mathbb{R}^3$ represents the contact force at the $i$-th contact point, and $\mathbf{J}_i(\mathbf{q}) \in \mathbb{R}^{3 \times (6+n)}$ is the associated contact Jacobian; here, we model collisions between rigid bodies on corresponding points and planes.
The contact impulse $ \boldsymbol{\lambda_i} = \Delta t\, \mathbf{f}_i$ is computed implicitly using time-stepping formulation based on a rigid contact model with anisotropic friction \cite{MPCC}. The contact constraints are expressed with a nonlinear friction cone and enforced via a smoothed complementarity condition \cite{TaoCIMPC} using a log-barrier relaxation, akin to the treatment in interior-point methods.

\subsection{Inertia Parameterization}
The robot dynamics are encoded through the inertial parameters of each link. The Lagrangian dynamics \eqref{eq:discrete_dynamics} is linear in a suitable inertial parameterization~\cite{wensing2024geometric}:
$
\mathbf{Y}(\mathbf{q},\mathbf{v},\mathbf{v}^+) \boldsymbol{\pi} = \boldsymbol{\tau},
$
where $\mathbf{Y}$ is the classical regressor, and $\boldsymbol{\pi}$ collects the inertial parameters of each link and can be written as: 
\begin{equation}
\boldsymbol{\pi}
= \bigl[m \ h_x \ h_y \ h_z \ I_{xx} \ I_{yy} \ I_{zz} \ I_{xy} \ I_{yz} \ I_{xz} \bigr]^{\top}
\in \mathbb{R}^{10}.
\end{equation}
Here, $m$ is the link mass, $\mathbf{h} = m\mathbf{c}$ is the first mass moment with $\mathbf{c}$ being the center-of-mass (COM) position expressed in the link frame. The remaining terms $I_\square$ compose the rotational inertia matrix $\mathbf{I} \in\mathbb{R}^{3 \times 3}$, which is defined at the link-fixed reference point expressed in the body frame. To enforce physical consistency of the inertia parameters, we formulate the pseudo-inertia matrix~\cite{wensing2024geometric} as:
\begin{equation}
    \mathcal{I} = \begin{bmatrix} \boldsymbol{\Sigma} & \mathbf{h} \\ \mathbf{h}^\intercal & m\end{bmatrix},
\end{equation}
and require it to be positive definite, where $\boldsymbol{\Sigma} = \frac{1}{2}\operatorname{Tr}(\mathbf{I}){1}_3 - \mathbf{I}$.
In nonlinear optimization, we employ the unconstrained Log-Cholesky inertial parameterization~\cite{LogParam}, parameterized by $\boldsymbol{\theta}$:
\begin{align}
  \mathcal{I} &= \mathbf{U} \mathbf{U} ^\intercal,  \quad
  \mathbf{U}
= e^{\alpha}
\begin{bmatrix}
e^{d_1} & s_{12} & s_{13} & t_1 \\
0       & e^{d_2} & s_{23} & t_2 \\
0       & 0       & e^{d_3} & t_3 \\
0       & 0       & 0       & 1
\end{bmatrix}, \\
\boldsymbol{\theta} &= \bigl[ \alpha \ d_1 \ d_2 \ d_3 \ s_{12} \ s_{23} \ s_{13} \ t_1 \ t_2 \ t_3 \bigr]^\intercal \in \mathbb{R}^{10}. 
\end{align}
This defines a smooth mapping from $\boldsymbol{\theta}$ to the standard inertial parameters $\boldsymbol{\pi}$, with a closed-form Jacobian $\frac{\partial \boldsymbol{\pi}}{\partial \boldsymbol{\theta}}$~\cite{LogParam}.

\subsection{Process Modeling with Contact Dynamics}
In time-stepping contact dynamics, physical consistency imposes additional constraints that couple the next-step configuration $\mathbf{q}^{+}$, velocity $\mathbf{v}^{+}$, and contact impulse $\boldsymbol{\lambda}$.

\noindent\textbf{Complementarity Condition:}
At the next time step $(\cdot)^{+}$, the signed distance $\phi(\mathbf{q}^{+})$ between two bodies and the normal contact impulse $\lambda^{n}$ over the time interval must be nonnegative. Moreover, the normal contact force can only be nonzero when the bodies are in contact. These conditions yield the classical complementarity condition:
\begin{equation}
    \phi(\mathbf{q}^{+}) \ge 0 \;\perp\; \lambda^{n} \ge 0.
    \label{eq:nonpenetration}
\end{equation}

\noindent\textbf{Maximum dissipation:}
The maximum-dissipation principle selects tangential friction forces that maximize kinetic-energy dissipation. For a single contact $i$, this can be written as
\begin{align}
    \min_{\boldsymbol{\lambda}_i} \quad & (\boldsymbol{\lambda}_i)^{\top}\,\mathbf{v}_{i}^{+}
    \label{eq:max_disp}\\
    \text{s.t.}\quad & \|\boldsymbol{\lambda}_i^{t}\|_2 \le \mu\, \lambda_i^{n},
\end{align}
where $\mu$ is the coefficient of friction, $\boldsymbol{\lambda}_i^{t}$ is the tangential component of the contact impulse, and $\mathbf{v}_{i}^{+}$ denotes the next-step relative velocity at contact $i$, e.g., $\mathbf{v}_{i}^{+} = \mathbf{J}_{i}(\mathbf{q}^{+})\,\mathbf{v}^{+}$.

\noindent\textbf{Mathematical Program with Complementarity Constraints:} 
The maximum dissipation principle~\eqref{eq:max_disp}, together with the discrete-time dynamics~\eqref{eq:discrete_dynamics} and non-penetration constraint~\eqref{eq:nonpenetration}, defines a complementarity-based contact model formulated as a Mathematical Program with Complementarity Constraints (MPCC)~\cite{MPCC}. Solving the MPCC yields the next-step robot state, compactly written as:
\begin{equation}
  \mathbf{x}^{+} \leftarrow\textbf{Dyn}(\mathbf{x}, \mathbf{u}).
    \label{eq:contact_dynamics}
\end{equation}
%The resulting second-order cone couples unilateral non-penetration with Coulomb friction: the normal component enforces unilateral contact, while the tangential cross-section encodes the friction bound. The associated cone complementarity conditions are equivalent to imposing the non-penetration condition together with maximum dissipation. 
\noindent\textbf{Contact Dynamics as Convex Optimization:}
The complementarity constrained time-stepping dynamics can be reformulated as a convex problem using Anitescu’s relaxation~\cite{Anitescu}. This convexification, however, may admit nonphysical behavior, where tangential sliding can induce a nonzero normal separation (i.e., spurious lift-off) \cite{PangSim}. In locomotion, this artifact is often acceptable because contacts are typically firm and short-lived, and most interactions occur near the sticking regime. In particular, friction can be expressed either as a second-order impulse cone $\mathcal{K}$:
\begin{equation}
\mathcal{K}
=
\left\{
(\lambda_i^{n},\boldsymbol{\lambda}_i^{t})\in \mathbb{R}\times\mathbb{R}^{2}
\;:\;
\|\boldsymbol{\lambda}_i^{t}\|_{2} \le \mu\,\lambda_i^{n}
\right\},
\end{equation}
or, equivalently, by its dual cone in contact velocity $\mathcal{K}^{*}$:
\begin{equation}
\mathcal{K}^{*}
=
\left\{
(v_i^{n},\mathbf{v}_i^{t})\in \mathbb{R}\times\mathbb{R}^{2}
\;:\;
\|\mathbf{v}_i^{t}\|_{2} \le \frac{1}{\mu}\,v_i^{n}
\right\}.
\end{equation}
Here $v_i^{n}=\mathbf{J}_i^{n}\mathbf{v}^{+}$ and $\mathbf{v}_i^{t}=\mathbf{J}_i^{t}\mathbf{v}^{+}$ denote the post-impact contact velocity in the normal and tangential directions, respectively. The impulse cone encodes the Coulomb friction bound, while the dual velocity cone encodes the admissible post-impact velocity with unilateral separation. The Anitescu’s cone complementarity conditions are then expressed as:
\begin{align}
(\lambda_i^{n},\boldsymbol{\lambda}_i^{t}) &\in \mathcal{K}, \\
(v_i^{n},\mathbf{v}_i^{t}) &\in \mathcal{K}^{*}, \\
(v_i^{n},\mathbf{v}_i^{t})^{\top}(\lambda_i^{n},\boldsymbol{\lambda}_i^{t})&=0.
\end{align}
These conditions are equivalent to imposing the non-penetration condition together with maximum dissipation, and correspond to the KKT conditions of the following second-order cone program (SOCP):
\begin{align} 
\min_{\mathbf{v}^{+}} \quad & \frac{1}{2}\, \left\| \mathbf{v}^{+} - \mathbf{v}_{\text{free}}^{+} \right\|_{\mathbf{M}(\mathbf{q})}^{2}, \label{eq:SOCP}
\\  
\text{s.t.} \quad & \left(\frac{\phi_i}{\Delta t} + \mathbf{J}^{n}_i\mathbf{v}^{+}\right)^{2} \;\ge\; \mu^{2}\, \left\| \mathbf{J}^{t}_i\mathbf{v}^{+} \right\|_{2}^{2}, \label{eq:cone}
\end{align}
where $\mathbf{v}_{\text{free}}^{+}$ denotes the contact-free integrated velocity. This SOCP can be efficiently solved by interior-point methods \cite{Mosek}.

\noindent\textbf{Smoothing of contact dynamics:}
Although Anitescu’s relaxation yields a convex formulation, the resulting contact dynamics remains only piecewise smooth. This induces discontinuous (or ill-defined) gradients, which hinder the direct use of gradient-based optimization methods. Prior work has addressed this issue in different ways: \cite{FastCIMPC} performs local linearization of the contact dynamics around a pre-planned trajectory, while \cite{KAISTCIMPC} modifies the discontinuous analytic gradients by introducing relaxed conditions. More recently, \cite{TaoCIMPC} proposed a log-barrier relaxation based smoothing approach of the second-order cone constraints in~\eqref{eq:SOCP}, which converts the SOCP into an unconstrained convex optimization problem:
\begin{align}
\min_{\mathbf{v}^{+}} \quad 
& \frac{1}{2}\, \left\| \mathbf{v}^{+} - \mathbf{v}_{\text{free}}^{+} \right\|_{\mathbf{M}(\mathbf{q})}^{2} \nonumber \\
&- \sum_i \frac{1}{\kappa}\,\log\!\left(
\frac{\left(\phi_i/\Delta t + \mathbf{J}^{n}_i\mathbf{v}^{+}\right)^{2}}{\mu^{2}}
- \left\|\mathbf{J}^{t}_i\mathbf{v}^{+}\right\|_2^{2}
\right), \label{eq:contact_smoothed}
\end{align}
with $\kappa$ being the weight of the log-barrier relaxation. This smoothing yields well-behaved and informative gradients around transitions of contact modes, enabling the optimizer to explore motions that involve changes in the contact modes. In this paper, the smoothed contact dynamics are solved using a customized Newton solver for fast computation, warm-started with a feasibility initializer that projects the previous time-step solution onto the normal feasibility constraints.

% \noindent{\textbf{Remark:}} \textcolor{red}{
% Should I emphasize this}

\noindent\textbf{Approximation of Latent Impulses:}
Although the smoothed contact dynamics are cast as an unconstrained optimization problem, the contact impulses—interpretable as the Lagrange multipliers of the original friction-cone (SOC) constraints \eqref{eq:cone}—can still be recovered from the first-order optimality conditions. In particular, for each contact indexed by $i$, the normal and tangential impulses can be approximated as:
\begin{equation}
\lambda_i^{n}
=
\frac{2\left(\phi_i/\Delta t + \mathbf{J}^{n}_i \mathbf{v}^{+}\right)}
{\mu^{2}\,\kappa\, s_i},
\end{equation}
\begin{equation}
\boldsymbol{\lambda}_i^{t}
=
-\frac{2\,\mathbf{J}^{t}_i \mathbf{v}^{+}}
{\kappa\, s_i},
\end{equation}
where:
\begin{equation}
s_i
=
\frac{\left(\phi_i/\Delta t + \mathbf{J}^{n}_i \mathbf{v}^{+}\right)^{2}}{\mu^{2}}
-
\left\|\mathbf{J}^{t}_i \mathbf{v}^{+}\right\|_{2}^{2}.
\end{equation}
As a result, the impulse as a latent variable is computed from the estimated trajectory that serves as the impulse estimation.

We will denote the smoothed contact dynamics by $\mathcal{F}(\cdot)$ in what follows. For the estimation task, the control input \(\mathbf{u}\) is assumed to be known from joint torque sensors. We introduce an additive process disturbance \(\boldsymbol{\delta}^{\mathbf{x}} \in \mathbb{R}^{n+6}\), which includes a floating-base component that is penalized more heavily, and model it as zero-mean Gaussian noise, \(\boldsymbol{\delta}^{\mathbf{x}} \sim \mathcal{N}(\mathbf{0}, \mathbf{Cov}_{\mathbf{x}})\), where $\mathbf{Cov}_{\mathbf{x}}$ denotes the covariance matrix. With this, the process dynamics of the robot are then written as:
\begin{equation}
\mathbf{x}^+ \leftarrow \mathcal{F}(\mathbf{x}, \mathbf{u}) \oplus \boldsymbol{\delta}^{\mathbf{x}},
\label{eq:contact_dynamics_estimation}
\end{equation}
where $\oplus$ represents additive stochastic disturbances in the state propagation. \eqref{eq:contact_dynamics_estimation} models contact dynamics while incorporating process uncertainty, such as actuation noise and unmodeled effects, which are assumed to follow the Gaussian distribution.

\subsection{Measurement Modeling}
To complement the increased fidelity of the contact dynamics, we incorporate all available measurements at the kinematic and actuation levels that are either onboard or readily accessible. The multi-sensor measurement vector $\mathbf{y}$ is modeled as a function of the state $\mathbf{x}$ and the control input $\mathbf{u}$:
\begin{equation}
    \mathbf{y} = \textbf{Meas}(\mathbf{x}, \mathbf{u}) \oplus \boldsymbol{\delta}^{\mathbf{y}},
\end{equation}
where $\boldsymbol{\delta}^{\mathbf{y}} \sim \mathcal{N}(\mathbf{0}, \mathbf{Cov}_{\mathbf{y}})$ represents measurement uncertainty, modeled as zero-mean Gaussian noises. The measurement model fuses heterogeneous sensing modalities, including encoder-measured joint positions and velocities, floating-base measurements from the motion-capture system, and actuator torque measurements that are commonly available on modern legged robots, e.g., via Quasi-Direct-Drive actuation. The measurement model does not require direct contact force or torque measurements; these quantities are inferred through the contact dynamics.

%\section{Contact-Involved Optimization-based Estimation}

\section{PRIME via Optimization-based Estimation}
Now we present the problem formulation of PRIME via optimization, and then we present our solution method that solves the optimization problem robustly and efficiently.

\subsection{Optimization-based Estimation}
Given a history of sensor measurements, optimal state estimation can be cast as a Maximum A Posteriori (MAP) problem. Full Information Estimation (FIE) solves this MAP problem over the entire time window by minimizing the corresponding negative log-likelihood~\cite{FIE}, thereby producing optimal estimates that leverage all available measurements. The formulation incorporates the system dynamics (Section~II.C), measurement model (Section~II.D), and physical constraints:
\begin{align*} 
&\min_{\mathbf{x}_{[0,T]},\boldsymbol{\delta}_{[0,T]}} \quad \Gamma(\mathbf{x}_{0}) \\ 
& + \sum_{k = 0}^{T-1}||\boldsymbol{\delta}^{\mathbf{x}}_k||^2_{\text{Cov}_{\mathbf{x}}^{-1}} + \sum_{k=0}^T||\boldsymbol{\delta}^{\mathbf{y}}_k||^2_{\text{Cov}_{\mathbf{y}}^{-1}} &&\tag{FIE} \label{eq:FIE}\\ 
\text{s.t.} \quad &\mathbf{x}_{k+1} \leftarrow {\mathcal{F}}(\mathbf{x}_k, \mathbf{u}_k) \oplus \boldsymbol{\delta}^{\mathbf{x}}_k, \ \forall k \in \{0,...,T-1 \} \\ 
&\mathbf{y}_k = \textbf{Meas}(\mathbf{x}_k) \oplus \boldsymbol{\delta}^{\mathbf{y}}_k, \ \forall k \in \{0,...,T \}, \\
&\mathbf{0} \leq \textbf{Constr}(\mathbf{x}_k), \ \forall k \in \{0,...,T \},
\end{align*}
in which dynamics and measurement uncertainties are minimized. Here, the subscript $\square_k$ denotes the time index,  and $\square_{[0, T]}$ denotes the trajectories from the start of time 0 to the end of time $T$. \(\Gamma(\cdot)\) denotes the prior term that anchors the initial state of the trajectory, and \(\boldsymbol{\delta}^{\mathbf{x}}_k\) and \(\boldsymbol{\delta}^{\mathbf{y}}_k\) represent process and measurement noise, respectively, modeled as zero-mean Gaussian random variables with covariances \(\mathbf{Cov}_{\mathbf{x}}\) and \(\mathbf{Cov}_{\mathbf{y}}\).

% \begin{figure}
%     \centering
%     \setlength{\abovecaptionskip}{0pt}
%     % \includesvg[width=0.95\linewidth]{figure/FIE_explain.svg}
%     \includegraphics[width=0.95\linewidth]{figure/FIE_explain.png}
%     \vspace{-0pt}
%     \caption{Illustration of FIE, arrival cost $\Gamma$ and their relationship with Full Information Estimation, where $\hat{x}_{i|k}$, $\{i,k \in \mathbb{N}^+|\ 0 \leq i \leq k \leq T \}$ is the estimate at time index $i$, using measurements from time 0 to time $k$.}
%     \vspace{-20pt}
%     \label{FIE_illu}
% \end{figure}
\subsection{Parameter and State Estimation via FIE}
During the estimation process, system parameters such as linkage inertia can be estimated, in other words, identified jointly with the robot state. Accordingly, the parameter-dependent dynamics in \eqref{eq:contact_dynamics_estimation} are written as
\begin{equation}
\mathbf{x}^+ \leftarrow \mathcal{F}(\mathbf{x}, \mathbf{u}, \boldsymbol{\theta}) \oplus \boldsymbol{\delta}^{\mathbf{x}},
\end{equation}
where $\boldsymbol{\theta}$ denotes the inertial-parameter representation introduced in Section~II.B, including a subset of link inertial parameters targeted for identification.
Assuming a Gaussian prior on the initial parameter estimate,
$\boldsymbol{\theta} \sim \mathcal{N}(\hat{\boldsymbol{\theta}}, \mathbf{Cov}_{\boldsymbol{\theta}})$,
we reformulate the original estimation problem in \eqref{eq:FIE} to jointly estimate states and parameters, yielding the Parameter Full-Information Estimation (PFIE) formulation of PRIME:
\begin{align*}
&\min_{\mathbf{x}_{[0,T]},\boldsymbol{\delta}_{[0,T]},\boldsymbol{\theta}}  \quad ||\boldsymbol{\theta} - \hat{\boldsymbol{\theta}} ||^2_{\text{Cov}_{\boldsymbol{\theta}}^{-1}} + \Gamma(\mathbf{x}_{0}) 
\\ & +  \sum_{k = 0}^{T-1}||\boldsymbol{\delta}^{\mathbf{x}}_k||^2_{\text{Cov}_{\mathbf{x}}^{-1}}  + \sum_{k=0}^T||\boldsymbol{\delta}^{\mathbf{y}}_k||^2_{\text{Cov}_{\mathbf{y}}^{-1}} &&\tag{PFIE} \label{eq:PFIE}\\
\text{s.t.} \quad 
&\mathbf{x}_{k+1} \leftarrow \mathcal{F}(\mathbf{x}_k, \mathbf{u}_k, \boldsymbol{\theta}) \oplus \boldsymbol{\delta}^{\mathbf{x}}_k, \ \forall k \in \{0,...,T-1 \} \\
&\mathbf{y}_k = \textbf{Meas}(\mathbf{x}_k) \oplus \boldsymbol{\delta}^{\mathbf{y}}_k, \ \forall k \in \{0,...,T \}. \end{align*}
% By adopting the Log-Cholesky inertia parameterization, the inertial parameters remain physically consistent without explicitly imposing constraints on the parameters, i.e., $\mathbf{0} \leq \textbf{Constr}(\boldsymbol{\theta}).$
By adopting the Log-Cholesky inertia parameterization, the inertial parameters remain feasible without explicitly imposing constraints, i.e., $\mathbf{0} \leq \textbf{Constr}(\boldsymbol{\theta}).$

\begin{algorithm}[t]
\caption{PRIME via DDP}
\label{alg:PDDP}
\begin{algorithmic}
\Require Initial trajectory $\tilde{\mathbf{x}}_t$, $\tilde{\boldsymbol{\delta}}_t$, and parameters $\tilde{\boldsymbol{\theta}}$
\While{not converged}
\State \textbf{\underline{Backward pass:}}
    \begin{align*}
        \nabla_{\Bar{\mathbf{x}}}{V} &:= \nabla_{\bar{\mathbf{x}}} L_N, \quad \nabla_{\Bar{\mathbf{x}}}^2{V} := \nabla_{\bar{\mathbf{x}}}^2 L_N 
    \end{align*}

    \For{$i = N-1 \to 0$}
        \begin{align*}
            A &= \partial_{\Bar{\mathbf{x}}} \mathcal{F}(\tilde{\mathbf{x}},\tilde{\boldsymbol{\delta}},\tilde{\boldsymbol{\theta}}),\\
            B &= \partial_{\boldsymbol{\delta}} \mathcal{F}(\tilde{\mathbf{x}},\tilde{\boldsymbol{\delta}},\tilde{\boldsymbol{\theta}}),\\
           \nabla_{\Bar{\mathbf{x}}} Q &= \nabla_{\Bar{\mathbf{x}}} L + A^\intercal \nabla_{\Bar{\mathbf{x}}}{V}, \\
            \nabla_{\boldsymbol{\delta}} Q &= \nabla_{\boldsymbol{\delta}} L + B^\intercal \nabla_{\Bar{\mathbf{x}}}{V}, \\
            \nabla_{\Bar{\mathbf{x}}}^2 Q &= \nabla_{\Bar{\mathbf{x}}}^2 L + A^\intercal (\nabla_{\Bar{\mathbf{x}}}^2{V}) A, \\
            \nabla_{\boldsymbol{\delta}}^2 Q &= \nabla_{\boldsymbol{\delta}}^2 L + B^\intercal (\nabla_{\Bar{\mathbf{x}}}^2{V}) B, \\
            \nabla_{\boldsymbol{\delta} {\Bar{\mathbf{x}}}}  Q &= \nabla_{\boldsymbol{\delta} \Bar{\mathbf{x}}} L + B^\intercal (\nabla_{\Bar{\mathbf{x}}}^2{V}) A.    
        \end{align*}
        
        \State Choose $\eta_i > 0$ s.t. $    \nabla_{\boldsymbol{\delta}}^2 Q := \nabla_{\boldsymbol{\delta}}^2 Q + \eta_i I > 0.$
        % \begin{align*}
        %     \nabla_{\boldsymbol{\delta}}^2 Q := \nabla_{\boldsymbol{\delta}}^2 Q + \eta_i I > 0.
        % \end{align*}

        \State Compute:
        \begin{align*}
& k_i = -\nabla_{\boldsymbol{\delta}}^2 Q^{-1} \nabla_{\boldsymbol{\delta}} Q \\
&K_i = -\nabla_{\boldsymbol{\delta}}^2 Q^{-1} \nabla_{\boldsymbol{\delta} {\Bar{\mathbf{x}}}} Q,  
 \\
&  \nabla_{\Bar{\mathbf{x}}}{V} = \nabla_{\Bar{\mathbf{x}}} Q + K_i^\intercal \nabla_{\boldsymbol{\delta}} Q,\\ &\nabla_{\Bar{\mathbf{x}}}^2{V} = \nabla_{\bar{\mathbf{x}}}^2 Q + K_i^\intercal \nabla_{\boldsymbol{\delta} \Bar{\mathbf{x}}} Q.
        \end{align*}
    \EndFor
\State \textbf{\underline{Forward pass:}}
\State Choose $\eta > 0$, s.t. $
            \nabla_{\Bar{\mathbf{x}}}^2 V :=  \nabla_{\Bar{\mathbf{x}}}^2 V + \eta I > 0. $
        % \begin{align*}
        %     \nabla_{\Bar{\mathbf{x}}}^2 V :=  \nabla_{\Bar{\mathbf{x}}}^2 V + \eta I > 0.
        % \end{align*}
\State Compute: \begin{align*}
    \delta \Bar{\mathbf{x}}_0^* &= - [\nabla_{\Bar{\mathbf{x}}}^2 V]^{-1} \nabla_{\Bar{\mathbf{x}}} V, \\
    \Bar{\mathbf{x}}_0^{'} &= \Bar{\mathbf{x}}_0 + \alpha \delta \Bar{\mathbf{x}}_0^*, \quad V^{'} := 0. 
\end{align*}
    \For{$i = 0 \to N-1$}
    \begin{align*}
        \boldsymbol{\delta}_i^{'} &= \boldsymbol{\delta}_i + \alpha k_i + K_i \delta \Bar{\mathbf{x}}_i, \\
        \Bar{\mathbf{x}}_{i+1}^{'} &= \begin{bmatrix}
            \mathcal{F} (\mathbf{x}_i^{'}, \boldsymbol{\delta}_i^{'}, \boldsymbol{\theta}^{'}) \\
            \boldsymbol{\theta}^{'}
        \end{bmatrix}, \\
        \delta \Bar{\mathbf{x}}_{i+1} &= \Bar{\mathbf{x}}_{i+1}^{'} - \begin{bmatrix}
            \mathbf{x}_i\\
            \boldsymbol{\theta}
        \end{bmatrix},\\
        V^{'} &= V^{'} + L_i(\Bar{\mathbf{x}}_i^{'},\boldsymbol{\delta}_i^{'}).
    \end{align*}
    \EndFor
    \begin{equation*}
        V^{'} = V^{'} + L_N(\Bar{\mathbf{x}}_N^{'}).
    \end{equation*}
\EndWhile
\State \Return optimal estimates of the physically-consistent trajectories $\mathbf{x}_t$ and parameters $\boldsymbol{\theta}$

\end{algorithmic}
\end{algorithm}

\subsection{Optimal Estimation via DDP}
Both the \eqref{eq:PFIE} and \eqref{eq:FIE} can be solved using Differential Dynamic Programming (DDP), by leveraging the differential properties of the contact dynamics~\eqref{eq:contact_smoothed}, even in the presence of complementarity constraints introduced by contact dynamics. We illustrate the derivations for the \eqref{eq:FIE} problem first. By analyzing the Bellman equation of~\eqref{eq:FIE}, the value function can be recursively expressed as:
\begin{align}
    V(\mathbf{x}) &= \min_{\boldsymbol{\delta}} Q(\mathbf{x}, \boldsymbol{\delta}), \\
    Q(\mathbf{x}, \boldsymbol{\delta}) &= V^+(\mathbf{x}^+) + L(\mathbf{x}, \boldsymbol{\delta}),
\end{align}
where $V(\cdot)$, $Q(\cdot)$, and $L(\cdot)$ are the value function, the local Q-function, and the stage cost, respectively. In DDP, the model uncertainty $\boldsymbol{\delta}$ is computed to minimize the local second-order approximation of $Q(\cdot)$ at the current estimated trajectory $\hat{(\cdot)}$:
\begin{align}
    Q(\mathbf{x}, \boldsymbol{\delta}) &\approx Q(\hat{\mathbf{x}}, \hat{\boldsymbol{\delta}}) + \Delta Q, \\
    \Delta Q &= \frac{1}{2} \begin{bmatrix}
        1 \\ \delta \mathbf{x} \\ \delta \boldsymbol{\delta}
    \end{bmatrix}^\intercal \begin{bmatrix}
        0 & \nabla_{\mathbf{x}} Q^\intercal & \nabla_{\boldsymbol{\delta}} Q^\intercal \\
        \nabla_{\mathbf{x}} Q & \nabla_{\mathbf{x}}^2 Q & \nabla_{\mathbf{x} \boldsymbol{\delta}} Q \\
        \nabla_{\boldsymbol{\delta}} Q & \nabla_{\boldsymbol{\delta} \mathbf{x}} Q & \nabla_{\boldsymbol{\delta}}^2 Q
    \end{bmatrix} \begin{bmatrix}
        1 \\ \delta \mathbf{x} \\ \delta \boldsymbol{\delta}
    \end{bmatrix}.
\end{align}
Minimizing $\Delta Q$ w.r.t. $\delta \boldsymbol{\delta}$ yields the optimal uncertainty:
\begin{equation}
\delta \boldsymbol{\delta}^* = \mathbf{K} \cdot \delta \mathbf{x} + \alpha \mathbf{k}, 
\end{equation}
\begin{equation}
\mathbf{K} = -\nabla_{\boldsymbol{\delta}}^2 Q^{-1} \nabla_{\boldsymbol{\delta} {\mathbf{x}}} Q,  
\quad \mathbf{k} = -\nabla_{\boldsymbol{\delta}}^2 Q^{-1} \nabla_{\boldsymbol{\delta}} Q, \nonumber    
\end{equation}
where $\alpha$ is a chosen step size via Armijo backtracking line search. When applying DDP to state estimation, the initial state is assumed to follow a prior distribution $\Gamma(\cdot)$, as indicated in \eqref{eq:FIE}. During DDP, the optimal perturbation to the initial state $\delta \mathbf{x}_0$ is generally non-zero and is computed as:
\begin{equation}
    \delta \mathbf{x}_0^* = - \left[\nabla_{\mathbf{x}_0}^2 V \right]^{-1} \nabla_{\mathbf{x}_0} V.
\end{equation}
\noindent{\textbf{Remark:}} For parameter and state estimation in \eqref{eq:PFIE} , the robot dynamics is augmented with the static parameter $\boldsymbol{\theta}$ dynamics within the horizon, and is similarly updated based on the optimal condition of the initial node:
\begin{align}
    \Bar{\mathbf{x}}^+ = \begin{bmatrix}
        \mathbf{x}^+ \\ \boldsymbol{\theta}^+ 
    \end{bmatrix} \leftarrow \begin{bmatrix}
        \mathcal{F}(\mathbf{x}, \mathbf{u}, \boldsymbol{\theta}) \oplus \boldsymbol{\delta}^{\mathbf{x}} \\
        \boldsymbol{\theta}
    \end{bmatrix}.
\end{align}
 The implementation of our DDP is detailed in Algorithm \autoref{alg:PDDP}.

\subsection{Multiple Shooting Differential Dynamic Programming}
 To robustly solve the \eqref{eq:PFIE} problem, we employ a multiple shooting variant of DDP, Feasibility-driven differential dynamic programming (FDDP). Unlike classical DDP, which typically assumes a dynamically feasible rollout, FDDP allows initialization and iteration with state-control trajectories that violate the dynamics. It achieves this by modifying the backward pass to account for defect variables and by permitting controlled dynamics infeasibility during early forward passes, progressively driving these defects toward zero across iterations. This ability to handle dynamics infeasibility is particularly valuable for contact estimation, where high-fidelity contact models can be numerically fragile—even with smoothing—and may otherwise hinder convergence. A detailed algorithmic description can be found in \cite{FDDP}.

\subsection{Analytic Gradient of Contact Dynamics}
As (F)DDPs rely on local approximations of the dynamics, the relationship among the perturbations
$\delta \bar{\mathbf{x}}^{+}$, $\delta \bar{\mathbf{x}}$, and $\delta \boldsymbol{\delta}$ is governed by the
infinitesimal dynamics induced by \eqref{eq:contact_dynamics_estimation}, i.e.,
\begin{equation}
\delta \bar{\mathbf{x}}^{+} = \mathbf{A}\,\delta \bar{\mathbf{x}} + \mathbf{B}\,\delta \boldsymbol{\delta},
\end{equation}
where $\mathbf{A}=\partial_{\bar{\mathbf{x}}}\mathcal{F}(\bar{\mathbf{x}},\boldsymbol{\delta})$ and
$\mathbf{B}=\partial_{\boldsymbol{\delta}}\mathcal{F}(\bar{\mathbf{x}},\boldsymbol{\delta})$ are the linearized
dynamics matrices used in Algorithm \ref{alg:PDDP} iterations.
Since the smoothed contact dynamics are posed as an unconstrained convex optimization, these derivatives can be obtained analytically via sensitivity analysis and the implicit function theorem.

As an example, consider differentiation with respect to the parameter vector $\boldsymbol{\theta}$. The first-order optimality condition of \eqref{eq:contact_smoothed} can be written as:
\begin{equation}
\nabla_{\mathbf{v}^{+}}\,c(\mathbf{v}^{+},\boldsymbol{\theta})
\;\triangleq\;
\mathbf{g}(\mathbf{v}^{+},\boldsymbol{\theta})
\;=\;\mathbf{0},
\end{equation}
where $c(\cdot)$ denotes the objective and $\mathbf{g}(\cdot)$ denotes its first-order derivative with respect to $\mathbf{v}^{+}$. 
Linearizing the optimality condition to first order yields:
\begin{equation}
\frac{\partial \mathbf{g}(\mathbf{v}^{+},\boldsymbol{\theta})}{\partial \mathbf{v}^{+}}\,\delta \mathbf{v}^{+}
\;+\;
\frac{\partial \mathbf{g}(\mathbf{v}^{+},\boldsymbol{\theta})}{\partial \boldsymbol{\theta}}\,\delta \boldsymbol{\theta}
\;=\;\mathbf{0},
\end{equation}
which gives the sensitivity of the post-impact generalized velocity with respect to the parameters:
\begin{equation}
\frac{\partial \mathbf{v}^{+}}{\partial \boldsymbol{\theta}}
=
-
\left(
\frac{\partial \mathbf{g}(\mathbf{v}^{+},\boldsymbol{\theta})}{\partial \mathbf{v}^{+}}
\right)^{-1}
\frac{\partial \mathbf{g}(\mathbf{v}^{+},\boldsymbol{\theta})}{\partial \boldsymbol{\theta}}.
\end{equation} 
For further discussion of the properties of this gradient, see \cite{TaoCIMPC} and the Supplementary Material. These analytic gradients are the key for efficient computation to solve the \eqref{eq:PFIE}. % via FDDP. 

\section{Results}
We evaluate PRIME in both simulation and hardware of several legged robots to simultaneously estimate their motion and uncertain inertia properties. First, we develop a custom MATLAB simulator to validate PRIME on a hopping robot. In this simulator, planar contact dynamics are formulated and solved as a linear complementarity problem (LCP) without smoothing. The required derivatives are obtained using the MATLAB Symbolic Math Toolbox, and the resulting \eqref{eq:PFIE} is solved using an open-source DDP implementation~\cite{ILQG}. 

We then utilize off-the-shelf high-fidelity MuJoCo simulation to demonstrate that PRIME generalizes to more complex and high-dimensional systems, including both quadrupedal Go2 and humanoid robot G1. For the hardware experiments, we obtained the data in the motion capture system that has 12 Opti-track cameras with a combination of Prime 13 and Prime 22. In these experiments, the robots are controlled using open-source reinforcement-learning policies \cite{isaaclab}. 

PRIME is fully implemented in C++ for fast computation. Analytic kinematic and dynamic derivatives are computed with Pinocchio~\cite{carpentier2019pinocchio}; notably, Pinocchio also provides the kinematic Hessians required for our analytic differentiation. The smoothed contact dynamics are solved using a customized Newton solver, and the resulting \eqref{eq:PFIE} problem is solved using FDDP as implemented in Crocoddyl~\cite{mastalli2020crocoddyl}.

\subsection{Hopper Experiments}
\begin{figure}[b]
    \centering
    \vspace{-15pt}
    \includegraphics[width=1.01\linewidth]{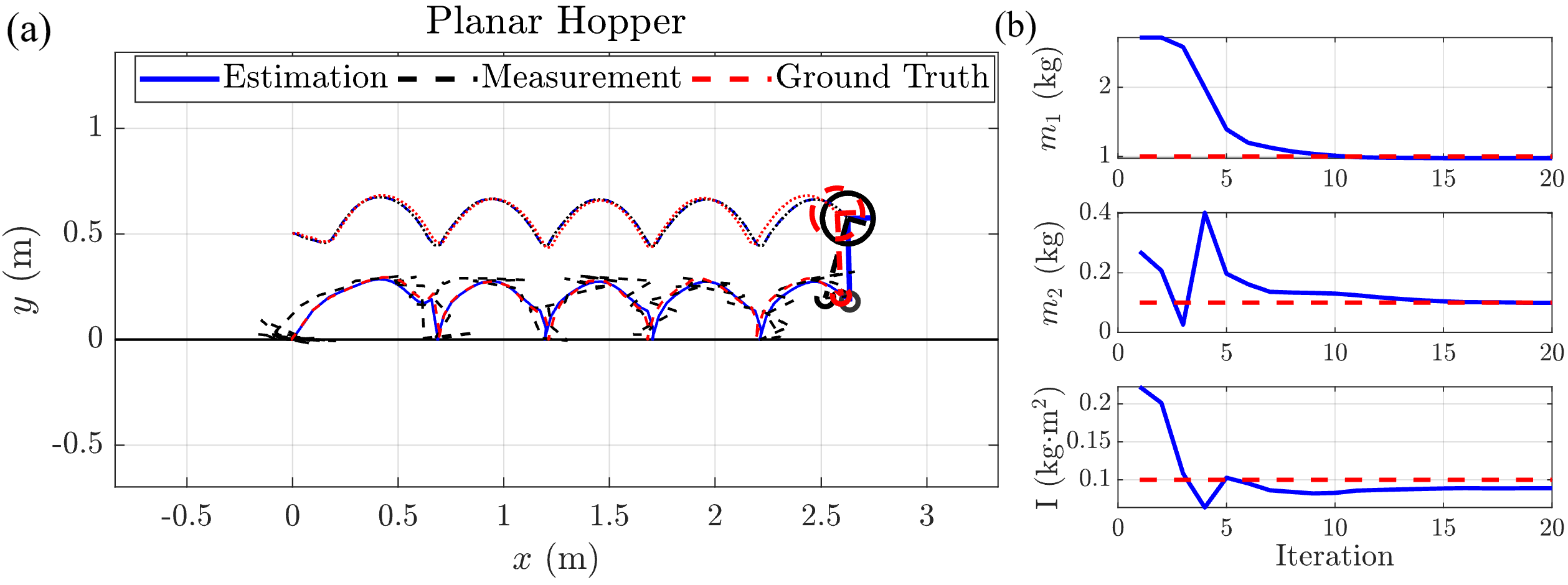}
    \vspace{-10pt}
    \caption{(a) Trajectories of the Hopper. (b) Convergence of the inertial parameters over iterations. The estimated inertial parameters converge to the ground-truth value, and the state trajectories align with the true trajectories with corrected contact modes.}
    \label{hopper_1}
    \vspace{-10pt}
\end{figure}

% \begin{figure}[t]
%     \centering
%     \setlength{\abovecaptionskip}{0pt}
%     \includegraphics[width=1.01\linewidth]{figure/hopper2.png}
%     \vspace{-6pt}
%     \caption{Hopper estimation results. From noise-corrupted measurements, the method reconstructs the state trajectory and contact forces, aligning closely with the simulated ground truth.}
%     \label{hopper_2}
%     \vspace{-0pt}
% \end{figure}

In the hopper experiments, we configure PFIE with a 2.5~s horizon and a time discretization of 0.025~s. The dataset is generated using a Contact-Implicit MPC controller~\cite{KAISTCIMPC}. Results on the PRIME hopping robot, as in Fig. \ref{hopper_1}, show that the proposed method is able to filter the noise-corrupted kinematic and actuator measurements to reconstruct the contact sequence and identify inertial parameters, starting from a biased initial inertia model. The iteration-wise convergence of the inertia estimates in Fig. \ref{hopper_1} further indicates consistent convergence toward the ground-truth values.

\subsection{PRIME for Quadrupedal Robots}
\begin{figure*}[t]
    \centering
    \setlength{\abovecaptionskip}{0pt}
    \includegraphics[width=0.98\textwidth]{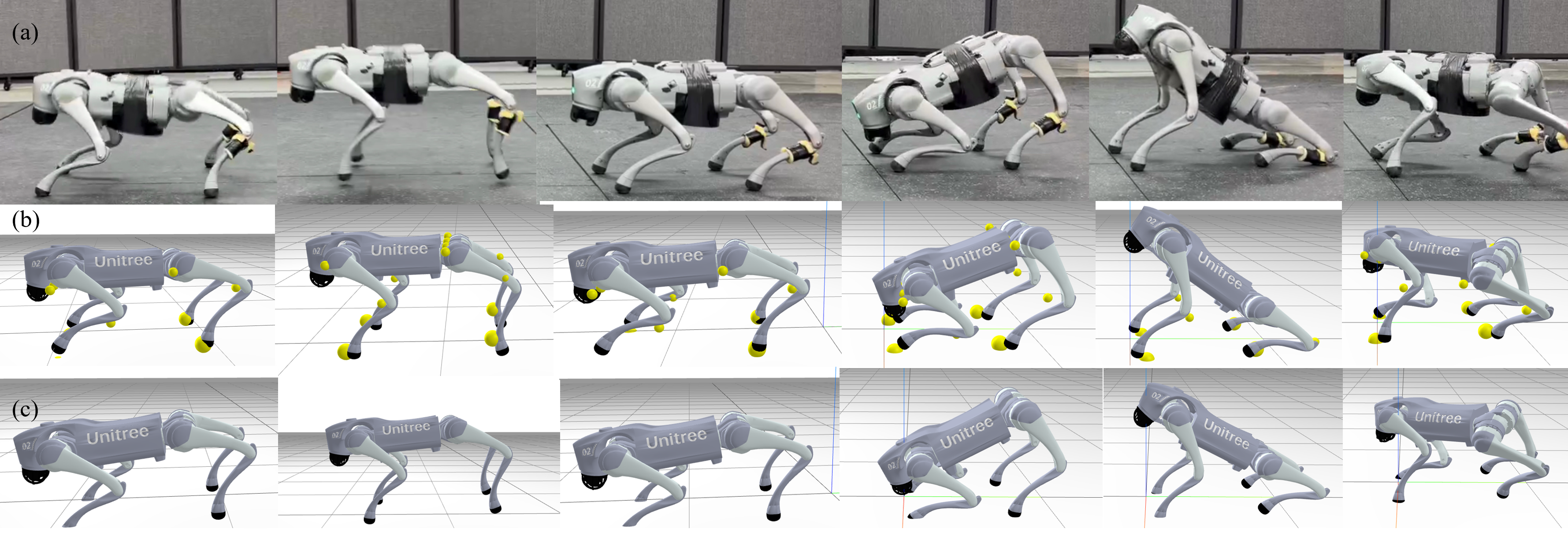}
    \vspace{-5pt}
    \caption{Quadrupedal robot Go2 motion estimation and inertia identification experiments with a 4.6 kg payload attached beneath the torso. (a) A 10 s segment comprising multiple locomotion behaviors, including torso rotations and forward hopping. (b) PRIME reconstructs the motion trajectory with dynamically consistent contact generated without using any contact measurements. The yellow spheres indicate the link poses from the raw kinematics. (c) The raw kinematics are physically inconsistent—showing non-physical contact penetrations—due to motion-capture orientation bias and jitters during locomotion.}
    \label{go2}
    \vspace{-10pt}
\end{figure*}

\begin{comment}
\begin{table*}[t]
\centering
\caption{Torso-link inertial properties identification experiments.}
\label{tab:torso_inertia_diag_compare}
\setlength{\tabcolsep}{8pt}
\renewcommand{\arraystretch}{1.15}
\begin{tabular}{lccccccc}
\toprule
\textbf{Case} &
$m$ [kg] &
$c_x$ [m] &
$c_y$ [m] &
$c_z$ [m] &
$I_{xx}$ &
$I_{yy}$ &
$I_{zz}$ \\
\midrule
Ground truth
& 6.927 & 0.022 & 0.000 & -0.005 & 0.025 & 0.098 & 0.108 \\
Simulation $m$ (+\textbf{3} kg)
& \textbf{9.975} & 0.034 & 0.138 & -0.012 & 0.033 & 0.138 & 0.150 \\
Simulation $c_z$ (- \textbf{0.1} m)
& 6.865 & 0.242 & 0.089 & \textbf{-0.105} & 0.024 & 0.089 & 0.097 \\
Real-world $m$ (+ \textbf{4.6} kg)
& \textbf{11.74} & 0.037 & 0.004 & \textbf{-0.013} & 0.014 & 0.274 & 0.274 \\
\bottomrule
\end{tabular}
\end{table*}
\end{comment}

\begin{figure*}[t]
    \centering
    \setlength{\abovecaptionskip}{0pt}
    \includegraphics[width=0.98\textwidth]{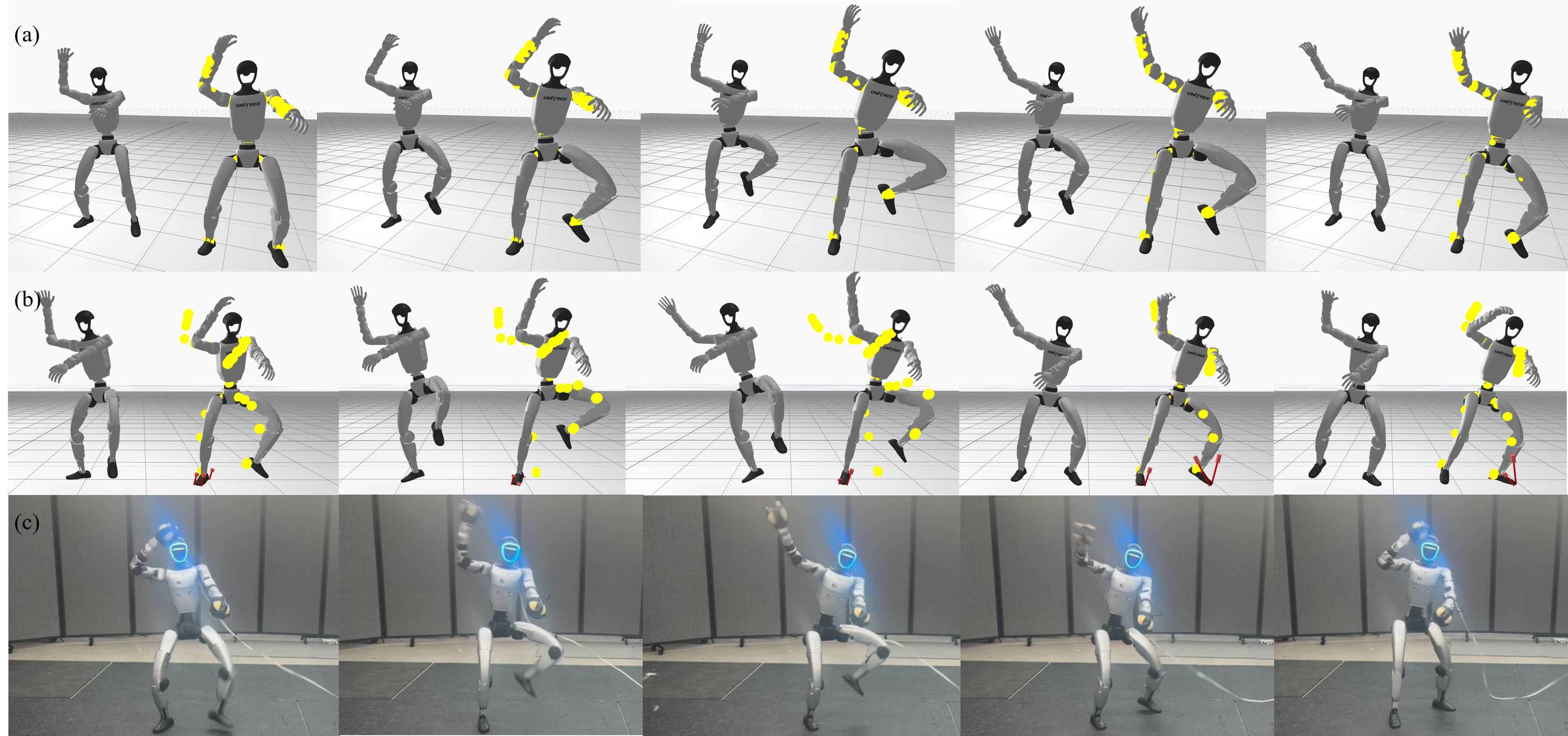}
    \vspace{-0pt}
    \caption{Humanoid G1 motion estimation experiment. (a) Simulation results showing the reconstructed trajectory closely aligned with the ground truth. (b,c) Real-robot results comparing raw kinematics (left) and PRIME reconstruction (right). Yellow spheres overlaid on the reconstructed motion indicate link positions from the raw kinematics, and red arrows denote the reconstructed contact forces at the foot corners. Despite noisy and contact-inconsistent kinematic measurements during the dancing sequence, PRIME reconstructs dynamically consistent motion with spatially distributed planar foot contact.}
    \label{humanoid_G1_results}
    \vspace{-20pt}
\end{figure*}
In the experiments on quadrupedal robots Go2, we jointly estimate robot motion and identify uncertain robot inertia in both simulation and hardware. Here, we assume the torso inertia is uncertain due to added payload onto the torso. In MuJoCo, we conduct two trials: (a) adding a 3~kg payload to the torso, and (b) shifting the torso center of mass downward by 0.1~m. In hardware, we attach two 2.3~kg barbell plates to the belly of the robot and then execute dynamic motions on the robot, including pronounced torso rotations and consecutive hopping with lateral velocity. For both simulation and hardware, PRIME is configured with a 10~s horizon and a 0.01~s time discretization (1000 samples). As summarized in Table~\ref{tab:torso_inertia_diag_compare}, the results show that the imposed inertial changes are accurately reflected in the estimated parameters; additionally, the reconstructed trajectories and contact forces closely match the ground truth. In the real-world experiments, PRIME similarly recovers the payload effect, estimating an added mass of approximately 4.8~kg and a downward shift of the torso COM consistent with the placement of the plates. The reconstructed state trajectory also remains consistent with the real experiments, even though the raw motion-capture kinematics exhibit orientation bias and jitter, which are clearly not contact- or dynamics-feasible—especially during the consecutive hopping behaviors. Despite these nonphysical measurement artifacts, PRIME still provides high-quality contact-consistent reconstructions as shown in Fig. \ref{go2}. Moreover, PRIME is quite efficient, converging within 200 s for a horizon of 1000 steps on a PC equipped with an Intel Core i9-13900 CPU.
% \balance
\begin{table}[t]
\centering
\caption{Identification of Torso Inertia and COM on Go2.}
\label{tab:torso_inertia_diag_compare}
\vspace{-5pt}
\begin{tabular}{lcccc}
\toprule
\textbf{Parameter} &
\textbf{Original} &
\textbf{Simulation} &
\textbf{Simulation } &
\textbf{Real-world} \\
\textbf{} &
\textbf{Values} &
\textbf{$m$ (+\textbf{3} kg)} &
\textbf{$c_z$ (-\textbf{0.1} m)} &
\textbf{$m$ (+\textbf{4.6} kg)} \\
\midrule
$m$ [kg]     & 6.927 & \underline{\textbf{9.975}} & 6.865 & \ul{\textbf{11.74}} \\
$c_x$ [m]   & 0.022 & 0.034 & 0.242 & 0.037 \\
$c_y$ [m]   & 0.000 & 0.138 & 0.089 & 0.004 \\
$c_z$ [m]   & -0.005 & -0.012 & \ul{\textbf{-0.105}} & \ul{\textbf{-0.013}} \\
$I_{xx}$    & 0.025 & 0.033 & 0.024 & 0.014 \\
$I_{yy}$    & 0.098 & 0.138 & 0.089 & 0.274 \\
$I_{zz}$    & 0.108 & 0.150 & 0.097 & 0.274 \\
\bottomrule
\end{tabular}
\vspace{-10pt}
\end{table}

\begin{table}[t]
\centering
\caption{Identification of Torso Inertia and COM on G1.}
\label{tab:com_mass_inertia_diag_compare_G1}
\vspace{-5pt}
\begin{tabular}{lcc}
\toprule
\textbf{Parameter} &
\textbf{Original} &
\textbf{Real-world} \\
\textbf{} &
\textbf{Values} &
\textbf{Total weight (+2.914kg)} \\
\midrule
$m$ [kg]     & 9.60     & \underline{\textbf{13.02}} \\
$c_x$ [m]    & 0.00332 & 0.03186 \\
$c_y$ [m]    & 0.00026 & -0.01649 \\
$c_z$ [m]    & 0.1798   & \underline{\textbf{0.1775}} \\
$I_{xx}$     & 0.1241   & 0.2669 \\
$I_{yy}$     & 0.1121   & 0.3075 \\
$I_{zz}$     & 0.0327  & 0.1097 \\
\bottomrule
\end{tabular}
\vspace{-15pt}
\end{table}

\begin{table}[t]
    \centering
    \caption{RMSEs and FIE Costs on G1 Hardware Experiments.}
    \label{tab:G1_RMSE}
    \vspace{-6pt}
    \begin{tabular}{|c|c|c|}
        \hline
        \textbf{Estimation Metric} & \textbf{With ID} & \textbf{W/O ID} \\
        \hline
        $\mathrm{RMSE}_{F}$ [N] & \textbf{24.486} & 26.141 \\
        \hline
        $\mathrm{Cost} [\times10^3]$ & \textbf{1.016} & 1.880 \\
        \hline
    \end{tabular}
    \vspace{-6pt}
\end{table}

\subsection{PRIME for Humanoids}
In the humanoid experiments, we collect diverse Unitree G1 locomotion behaviors, including walking, running, and dancing. Ground-truth contact forces for a dancing sequence are collected using a Bertec-4060 six-axis force plate (1000~Hz, $\pm 0.4$~N resolution). For all humanoid trials, PRIME uses a 15~s horizon and a time discretization of 0.01~s (1500 samples), sufficient to capture diverse motions and multiple contact transitions. PRIME solves each problem in approximately 400~s on the same PC. Planar foot contact is approximated by a four-point contact model at the corners of each foot, consistent with the rigid-foot design of the G1.

\begin{figure}[t]
    \centering
    \setlength{\abovecaptionskip}{0pt}
    \includegraphics[width=1.0\linewidth]{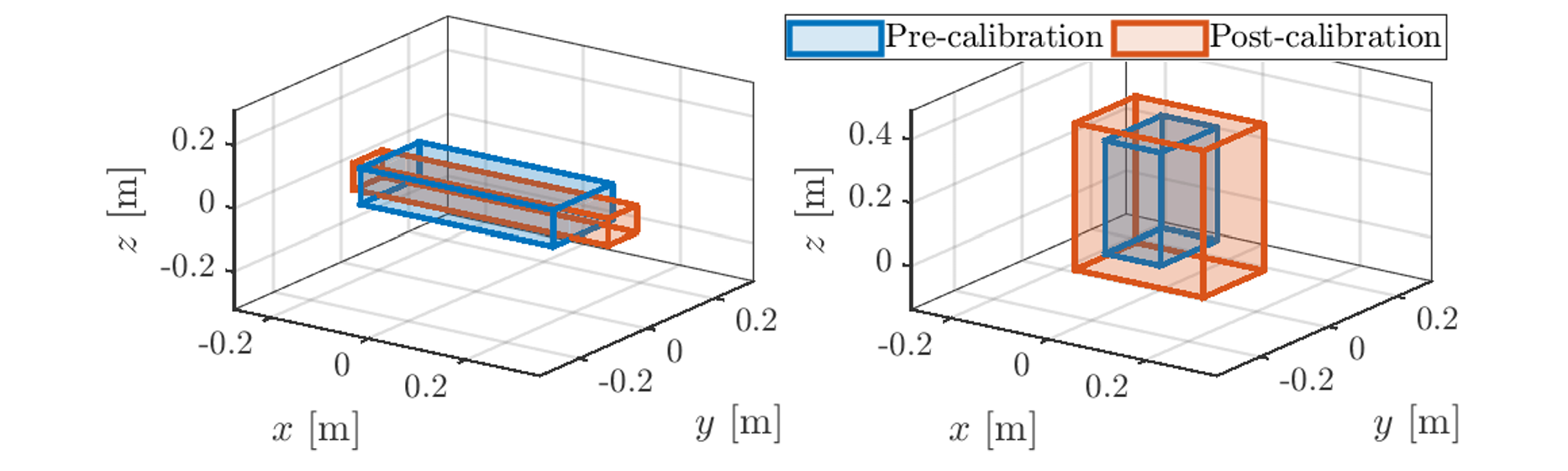}
    \vspace{-15pt}
    \caption{Equivalent inertia-box of hardware experiments (a) Go2; (b) G1.}
    \label{inertia_box}
    \vspace{-10pt}
\end{figure}
\begin{figure}[!t]
    \centering
    \setlength{\abovecaptionskip}{0pt}
    \includegraphics[width=0.95\linewidth]{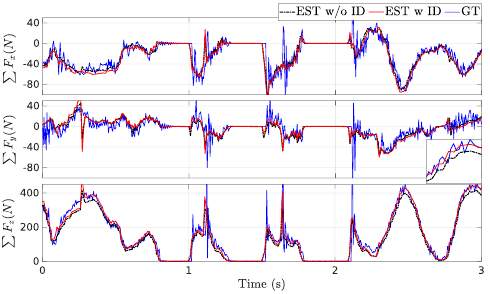}
    % \includesvg[width=0.95\linewidth]{figure/force_compare_3_sub.svg}
    \vspace{-5pt}
    \caption{Force comparison for the G1 humanoid (left foot) over a dancing motion. The identification module estimates inertial discrepancies and corrects the reconstructed contact forces, improving accuracy against the force-plate ground truth.}
    \label{G1_force_real}
    \vspace{-20pt}
\end{figure}
% \begin{figure}[t]
%     \centering
%     \setlength{\abovecaptionskip}{0pt}
%     \includegraphics[width=1.01\linewidth]{figure/G1_force_right.png}
%     \vspace{-6pt}
%     \caption{Force estimation of humanoid G1 (right foot) during a dancing motion. (a) PRIME reconstruction on the real hardware. (b) PRIME reconstruction in simulation, compared against ground-truth forces in MuJoCo. Because the dancing motion is highly dynamic, reliable ground-truth contact forces on hardware are difficult to acquire; instead, we use the same motion sequence in MuJoCo as an indirect reference. The simulated force value and contact sequence closely match those reconstructed on the real robot.}
%     \label{force_sum}
%     \vspace{-0pt}
% \end{figure}

% \begin{figure}[t]
%     \centering
%     \setlength{\abovecaptionskip}{0pt}
%     \includegraphics[width=1.01\linewidth]{figure/4pointforce.png}
%     \vspace{-6pt}
%     \caption{Humanoid G1 simulation force estimation using a four-point contact model on each foot. This four-point approximation accurately reconstructs planar foot contact, enabling high-fidelity contact reconstruction.}
%     \label{4point}
%     \vspace{-0pt}
% \end{figure}

% \begin{figure}[t]
%     \centering
%     \setlength{\abovecaptionskip}{0pt}
%     \includegraphics[width=1.01\linewidth]{figure/G1_sim.png}
%     \vspace{-6pt}
%     \caption{Unitree G1 trajectory estimation results in simulation.}
%     \label{G1_state_sim}
%     \vspace{-5pt}
% \end{figure}

Although the ground-truth inertial parameters of the G1 are unavailable, our experiments reveal a measurable discrepancy between the nominal model and the physical robot. The scale-measured total robot mass is $38.029$~kg, whereas the nominal model mass is $35.115$~kg, yielding a difference of $2.914$~kg. This discrepancy may be attributed to the battery, which weighs $2.496$~kg. As shown in Table~\ref{tab:com_mass_inertia_diag_compare_G1}, PRIME estimates an additional mass of approximately $3.4$~kg, while preserving a similar COM location and increasing the rotational inertia accordingly. The corresponding equivalent inertia boxes further suggest a reasonable localized load distribution, as illustrated in Fig.~\ref{inertia_box}. This identification result is also reflected in the contact-force comparison. With system identification enabled, PRIME achieves a lower cost in~\eqref{eq:FIE} and produces force estimates that more closely match the force-plate ground truth, with smaller measurement residuals and lower RMSEs, as shown in Fig.~\ref{G1_force_real} and Table~\ref{tab:G1_RMSE}. These results indicate that the improved inertial model leads to more accurate force reconstruction. PRIME further reconstructs smooth planar contact transitions, including heel-to-toe rolling and diverse contact patterns during dancing on a complex humanoid platform as illustrated in Fig. \ref{humanoid_G1_results}. To the best of our knowledge, this level of contact-force reconstruction has not been previously demonstrated for such humanoid motions. Additional analysis and comparisons with alternative contact-modeling choices are provided in the Supplementary Material.

% In the humanoid experiments, we collect a diverse set of Unitree G1 locomotion behaviors, including walking, running, and dancing. Because ground-reaction forces are difficult to measure under the complex, distributed contacts between the planar feet and the ground, we instead report a simulation-based comparison of the estimated contact forces. For all humanoid trials, we configure PRIME with a 15~s horizon and a time discretization of 0.01~s (1500 samples), which is long enough to capture diverse motions and multiple contact transitions. PRIME solves the problem within around 400 s of computation on the same PC. Planar foot contact is approximated using a four-point contact model at the corners of each foot, consistent with the rigid-foot design of G1.

% Fig. \ref{humanoid_G1_results} shows that, with smoothed contact dynamics and the resulting informative gradients, the algorithm leverages multiple shooting to converge robustly to consistent objective values across all datasets, despite the numerical challenges posed by frequent contact transitions, long horizons, and noisy measurements. As shown in Fig. \ref{force_sum}, \ref{4point}, PRIME reliably reconstructs planar contact with accurate, smooth transitions, capturing effects such as heel-to-toe roll and the diverse contact patterns that arise during dancing on a complex humanoid platform—performance that, to the best of our knowledge, has not previously been demonstrated. More extensive results and analysis can be found in the Supplementary Material.

\section{Conclusion and Future Work}
This paper presented PRIME, an optimization-based estimation framework that refines measured kinematics into dynamically consistent motion trajectories by explicitly reasoning about contact dynamics and inertial properties. By formulating motion and parameter estimation as a Maximum A Posteriori problem with differentiable contact dynamics, PRIME jointly reconstructs state trajectories, frictional contact forces, and physically consistent inertial parameters. Experimental results on diverse legged robot hardware demonstrate that PRIME produces contact-consistent motion reconstructions and accurate inertial identification over long horizons, even when raw kinematic measurements are noisy or physically inconsistent. These results highlight the importance of enforcing physical consistency for reliable motion estimation in contact-rich robotic systems.

While this work focuses on legged and humanoid locomotion, the underlying formulation of PRIME is not specific to locomotion and naturally extends to other contact-rich behaviors, including robotic manipulation and loco-manipulation, where intermittent contacts and unobserved interaction forces pose similar challenges. An important direction for future work is to study the observability and identifiability of motion, contact forces, and inertial parameters under limited sensory information, and to understand how task structure and contact excitation influence estimation performance. In addition, by transforming raw robot logs into force- and contact-consistent motion reconstructions, PRIME suggests a pathway for enriching real-robot datasets with physically grounded supervision, which could support downstream data-driven approaches, including recent vision–language–action and large-scale behavior models.

% offering a physics-based mechanism to enrich the dimensionality and physical fidelity of real-robot datasets, which could support emerging robot foundation and vision–language–action models that depend on large-scale, high-quality interaction data.

% …which may support downstream data-driven approaches, including recent vision–language–action and large-scale behavior models, by providing physically grounded motion and interaction supervision from real robots.

%%%%%%%%%%%%%%%%%%%%%%%%%%%%%%%%%%%%%%%%%%%%%%%%%%%%%%%%%%%%%%%%%%%%%%%%%%%%%%%%

%%%%%%%%%%%%%%%%%%%%%%%%%%%%%%%%%%%%%%%%%%%%%%%%%%%%%%%%%%%%%%%%%%%%%%%%%%%%%%%%

%%%%%%%%%%%%%%%%%%%%%%%%%%%%%%%%%%%%%%%%%%%%%%%%%%%%%%%%%%%%%%%%%%%%%%%%%%%%%%%%

%%%%%%%%%%%%%%%%%%%%%%%%%%%%%%%%%%%%%%%%%%%%%%%%%%%%%%%%%%%%%%%%%%%%%%%%%%%%%%%%
\newpage
% \addtolength{\textheight}{-5.0cm}
\bibliographystyle{IEEEtran}
\bibliography{references}

% \newpage

\section{Supplementary Materials}
\begin{figure}[!htbp]
    \centering
    \setlength{\abovecaptionskip}{0pt}
    \includegraphics[width=1.01\linewidth]{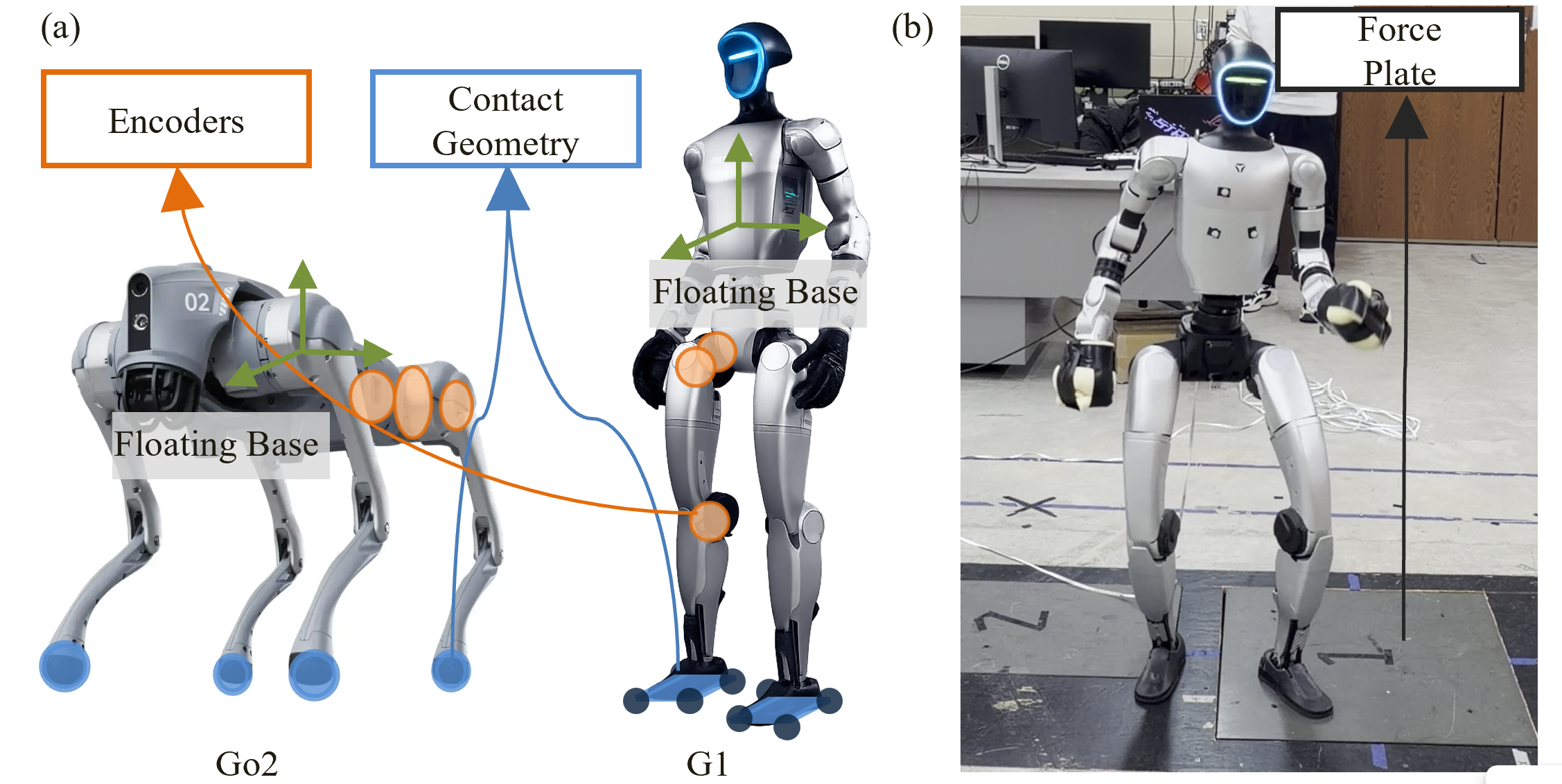}
    \vspace{-6pt}
    \caption{(a) Robot experimental configurations for Unitree Go2 and G1.
Point contact model is used to approximate the underlying contact geometry, which makes the framework invariant under point- or planar-contact for both robots. (b) G1 experimental setup with contact force ground truth from force plate.}
    \label{fig:config}
    \vspace{-10pt}
\end{figure}

The supplementary materials present additional results and analysis for the proposed PRIME framework, including multimedia demonstrations, supplementary plots, and extended experimental evaluations. All experiments are conducted on the Unitree Go2 and G1 robots, as illustrated in Fig.~\ref{fig:config}. %The code will be released after the double-blind review process.
\subsection{Multimedia Materials}
Table~\ref{tab:multimedia} summarizes the multimedia materials used to visualize the reconstructed motion and contact force trajectories for experiments on both Unitree Go2 and G1.
\begin{table}[!htbp]
\centering
\caption{Index to multimedia materials.}
\label{tab:multimedia}
\begin{tabular}{@{} c c p{0.55\linewidth} @{}}
\toprule
\textbf{Extension} & \textbf{Media Type} & \textbf{Description} \\
\midrule
1  & Video & Hardware results on Unitree Go2 with a 4.6\,kg payload attached to the belly, demonstrating torso rotations and forward hopping. \\
2  & Video & Hardware results on Unitree G1, demonstrating reconstructed motion for a dancing sequence. \\
3  & Video & Hardware results on Unitree G1 during force-plate data collection, demonstrating reconstructed motion for an additional dancing sequence.\\
4  & Video & Simulation results on Unitree Go2 with a 3\,kg payload added to the torso, demonstrating torso rotations and forward hopping. \\
5  & Video & Simulation results on Unitree G1, demonstrating reconstructed motion for a similar dancing sequence as 2. \\
\bottomrule
\end{tabular}
\end{table}
% \begin{table}[!htbp]
% \centering
% \caption{Index to multimedia materials.}
% \label{tab:multimedia}
% \begin{tabular}{@{} c c p{0.55\linewidth} @{}}
% \toprule
% \textbf{Extension} & \textbf{Media Type} & \textbf{Description} \\
% \midrule
% 1  & Video & Simulation results on Unitree Go2 with a 3\,kg payload added to the torso, demonstrating torso rotations and forward hopping. \\
% 2  & Video & Simulation results on Unitree Go2 with the center of mass shifted downward by 0.1\,m, demonstrating torso rotations and forward hopping. \\
% 3  & Video & Hardware results on Unitree Go2 with a 4.6\,kg payload attached to the belly, demonstrating torso rotations and forward hopping. \\
% 4  & Video & Simulation results on Unitree G1, demonstrating reconstructed motion for a dancing sequence. \\
% 5  & Video & Hardware results on Unitree G1, demonstrating reconstructed motion for the same dancing sequence as 4. \\
% 6  & Video & Hardware results on Unitree G1 during force-plate data collection, demonstrating reconstructed motion for an additional dancing sequence.\\
% \bottomrule
% \end{tabular}
% \end{table}
\subsection{Weight Setting}
Table \ref{tab:weight} summarizes the default weight settings used in PRIME. These weights are chosen intuitively based on the observed measurement quality. In particular, the floating-base linear velocity is obtained by differentiating the mocap position measurements and is therefore more noise-sensitive, so it is assigned a lower weight. Moreover, due to calibration bias, motion jitter, and marker occlusion, the mocap-based orientation measurements are of degraded quality, whereas the angular-velocity measurements from the IMU are more robust. To mitigate the effect of noisy data, we include a regularization term that penalizes deviations of the inertial parameters from their nominal values provided by the manufacturer.
\subsection{PRIME for Quadrupedal Robots}

Additional figures are provided for the hardware experiments on the quadrupedal robot Go2 in Section V.B, where reconstructed trajectories are compared against dynamically infeasible raw kinematic measurements (Figs.~\ref{fig:Go2_base}-\ref{fig:Go2_tau}). The roll-direction discrepancy in Fig.~\ref{fig:Go2_base} and Supplementary Video~1 illustrates a common practical error arising from imperfect mocap calibration. More broadly, even with careful calibration, mocap remains a purely kinematic measurement system and does not enforce dynamic or contact consistency by construction; this can lead to physically inconsistent contact states, such as artificial foot-ground separation or penetration, as shown in Fig.~3(c).

% The orientation calibration error of the mocap system is evident in the roll direction in Fig.~\ref{fig:Go2_base} and Supplementary Video~1.
\subsection{PRIME for Humanoid Robots}
% \begin{figure}[t]
%     \centering
%     \setlength{\abovecaptionskip}{0pt}
%     \includegraphics[width=1.01\linewidth]{figure/G1_state_real_dance.png}
%     \vspace{-6pt}
%     \caption{State estimation of humanoid G1 during a dancing motion.}
%     \label{G1_state_real_dance}
%     \vspace{-0pt}
% \end{figure}

% Requires: \usepackage{booktabs}
% Optional for math: \usepackage{amsmath,amssymb}
% \begin{figure}[t]
%     \centering
%     \setlength{\abovecaptionskip}{0pt}
%     % \includegraphics[width=1.0\linewidth]{figure/force_compare_3_sub.svg}
%     \includesvg[width=1.0\linewidth]{figure/force_compare_3_sub.svg}
%     \vspace{-10pt}
%     \caption{Force estimation comparison for the G1 humanoid (left foot) over a dancing motion. The identification module estimates inertial discrepancies and corrects the reconstructed contact forces, improving accuracy against the force-plate ground truth.}
%     \label{G1_force_real}
%     \vspace{-0pt}
% \end{figure}

\begin{table}[t]
\centering
\caption{Default cost weights used by PRIME in the Go2 and G1 experiments.}
\label{tab:weight}
\begin{tabular}{@{} l c c @{}}
\toprule
\textbf{Cost Weights} & \textbf{Expression} & \textbf{Coefficient} \\
\midrule
\multicolumn{3}{@{}l}{\textbf{Floating base cost}} \\
Position noise          & $\|p - \tilde{p}\|^2$                 & $4\times10^2$ \\
Velocity noise          & $\|v - \tilde{v}\|^2$                 & $1\times 10^1$ \\
Orientation noise        & $\|log(R\tilde{R}^\intercal)\|^2$ & $3\times 10^1$\\
Angular velocity noise  & $\|\omega - \tilde{\omega}\|^2$       & $1.5\times 10^2$ \\
\midrule
\multicolumn{3}{@{}l}{\textbf{Joint cost}} \\
Position noise    & $\|\alpha - \tilde{\alpha}\|^2$     & $2\times 10^2$ \\
Velocity noise    & $\|\dot{\alpha} - \dot{\tilde{\alpha}}\|^2$ & $4\times 10^1$ \\
Torque noise      & $\|\tau - \tilde{\tau}\|^2$            & $2\times 10^1$ \\
\midrule
\multicolumn{3}{@{}l}{\textbf{Parameter regularization}} \\
Log-Cholesky            & $\|\theta - \theta_0\|^2$                               & $4 \times10^{-2}$ \\
\bottomrule
\end{tabular}
\end{table}

\begin{figure}[t]
    \centering
    \setlength{\abovecaptionskip}{0pt}
    \includegraphics[width=1.01\linewidth]{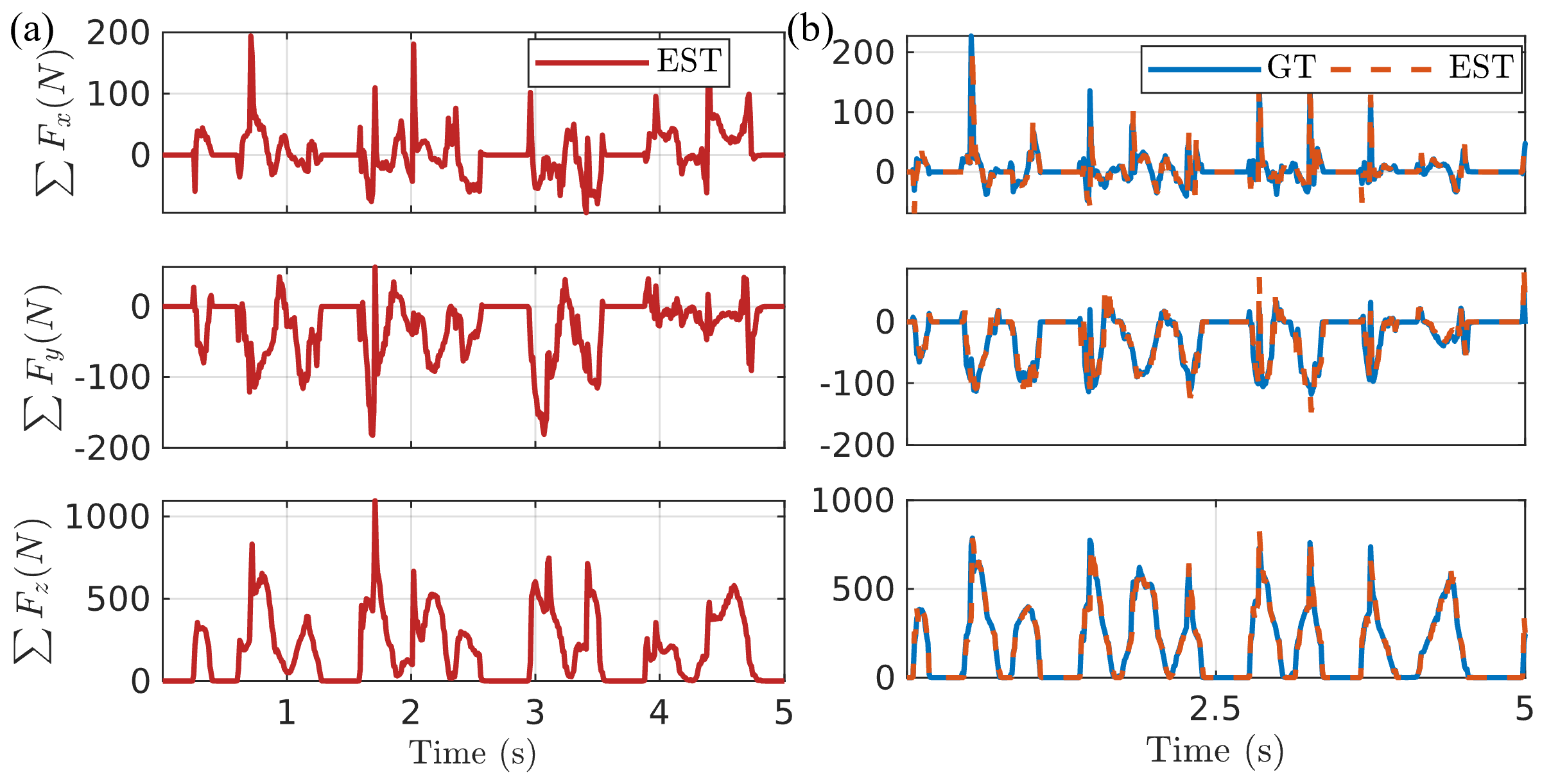}
    \vspace{-15pt}
    \caption{Force estimation for the G1 humanoid right foot during a dancing motion. (a) PRIME reconstruction on real hardware. (b) PRIME reconstruction in simulation, compared against MuJoCo ground-truth forces. The simulated force profiles and contact sequence closely match those reconstructed from the real robot under the same motion, suggesting that the chosen contact geometry provides a consistent representation across simulation and hardware.}
    \label{force_sum}
    \vspace{-10pt}
\end{figure}

\begin{figure}[t]
    \centering
    \setlength{\abovecaptionskip}{0pt}
    \includegraphics[width=1.0\linewidth]{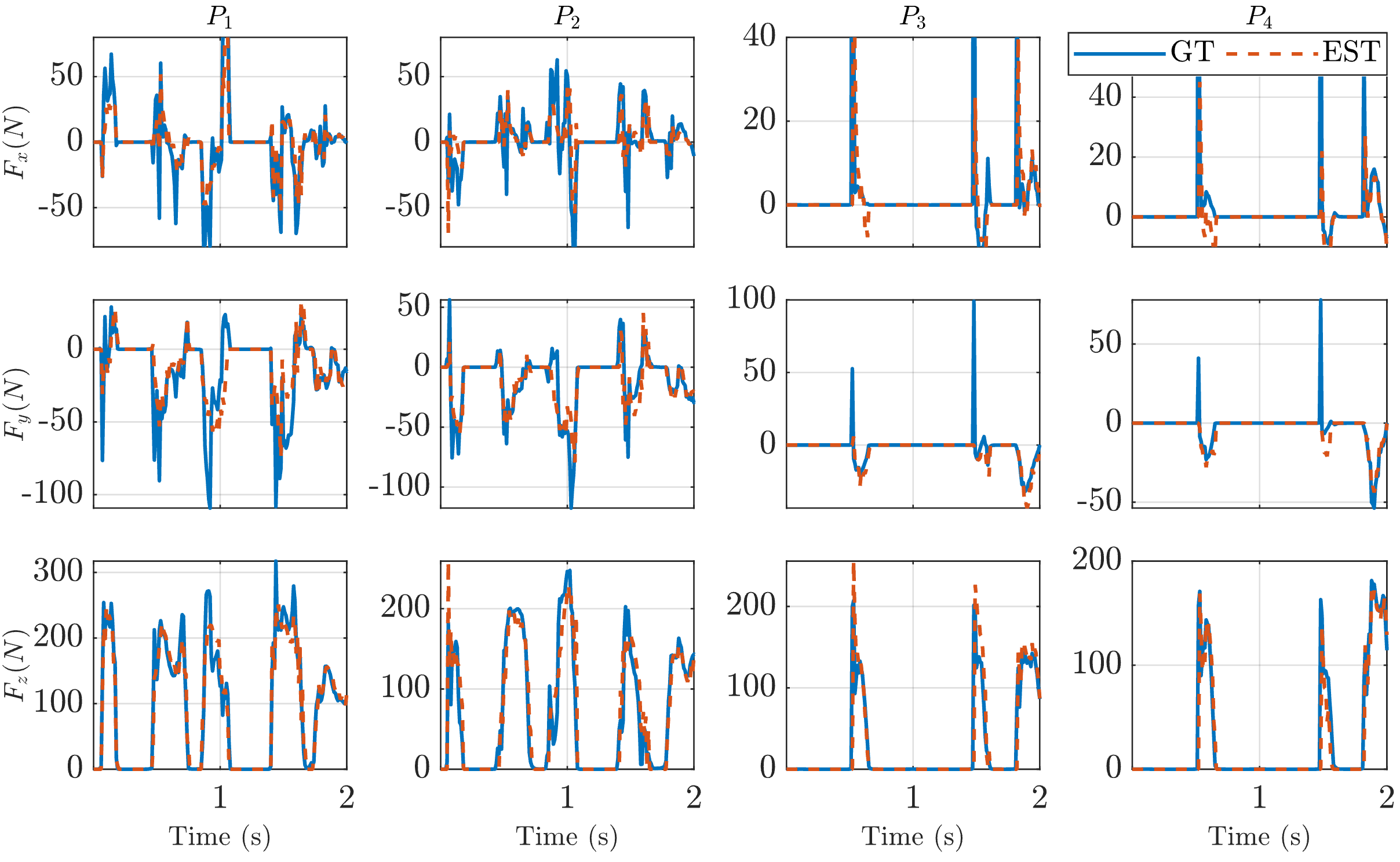}
    \vspace{-15pt}
    \caption{Humanoid G1 simulation force estimation using a four-point contact model on each foot. This four-point approximation accurately reconstructs planar foot contact, enabling high-fidelity contact reconstruction.}
    \label{4point}
    \vspace{-15pt}
\end{figure}

Additional figures are provided here to complement the G1 humanoid experiments in Section~V.C and provide a more complete view of the reconstructed motion and contact behavior. Since ground-truth contact forces at individual foot corners are unavailable in the real-world experiments, we further assess the reconstructed contact forces using simulation ground truth and by checking consistency between hardware and simulation. Fig.~\ref{force_sum} compares the reconstructed right-foot forces for the same dancing motion sequence in hardware and simulation. The force profiles and contact sequence are consistent across hardware and simulation, suggesting that the chosen contact geometry, which approximates planar foot contact using corner points, provides a reasonable representation in both simulation and hardware. Fig.~\ref{4point} shows the per-corner force reconstruction in simulation, where the estimated contact schedule and force profiles closely match the ground truth.

To further assess the reconstructed motion, Figs.~\ref{fig:G1_base}-\ref{fig:G1_tau} compare the estimated trajectories with raw kinematic measurements. The reconstructed kinematic trajectories align with the motion-capture measurements and raw kinematics while correcting orientation misalignment and reducing jitter. Following the identification analysis in Section~V.C, the torque reconstruction also aligns with the sensor readings after correcting the torso inertia. The remaining torque mismatches in Fig.~\ref{fig:G1_tau} occur primarily at the upper-limb joints, where the three-finger hands are not modeled.
% In the humanoid experiments, we collect a diverse set of Unitree G1 locomotion behaviors, including walking, running, and dancing. Because ground-reaction forces are difficult to measure under the complex, distributed contacts between the planar feet and the ground, we instead report a simulation-based comparison of the estimated contact forces. For all humanoid trials, we configure PRIME with a 15~s horizon and a time discretization of 0.01~s (1500 samples), which is long enough to capture diverse motions and multiple contact transitions. PRIME solves the problem within around 400 s of computation on the same PC. Planar foot contact is approximated using a four-point contact model at the corners of each foot, consistent with the rigid-foot design of G1.
% Fig. \ref{humanoid_G1_results} shows that, with smoothed contact dynamics and the resulting informative gradients, the algorithm leverages multiple shooting to converge robustly to consistent objective values across all datasets, despite the numerical challenges posed by frequent contact transitions, long horizons, and noisy measurements. As shown in Fig. \ref{force_sum}, \ref{4point}, PRIME reliably reconstructs planar contact with accurate, smooth transitions, capturing effects such as heel-to-toe roll and the diverse contact patterns that arise during dancing on a complex humanoid platform—performance that, to the best of our knowledge, has not previously been demonstrated. More extensive results and analysis can be found in the Supplementary Material.

\begin{table}[t]
    \centering
    \caption{Reconstruction errors and FIE cost in G1 simulation.}
    \vspace{-10pt}
    \label{tab:G1_baseline_RMSE}
    \begin{tabular}{|c|c|c|c|}
        \hline
        \textbf{Method} & $\mathbf{RMSE}_{F}$ [N] & $\mathbf{Realtive Error}_{F}$ [\%] & \textbf{Cost} [$\times 10^3$] \\
        \hline
        PRIME & \textbf{19.833} & \textbf{10.494} & \textbf{1.127} \\
        \hline
        Baseline& 73.103 & 52.693 & 1.36 \\
        \hline
    \end{tabular}
    \vspace{-10pt}
\end{table}
\begin{figure}[!t]
    \centering
    \setlength{\abovecaptionskip}{0pt}
    % \includesvg[width=1.0\linewidth]{figure/baseline_compare.svg}
    \includegraphics[width=1.0\linewidth]{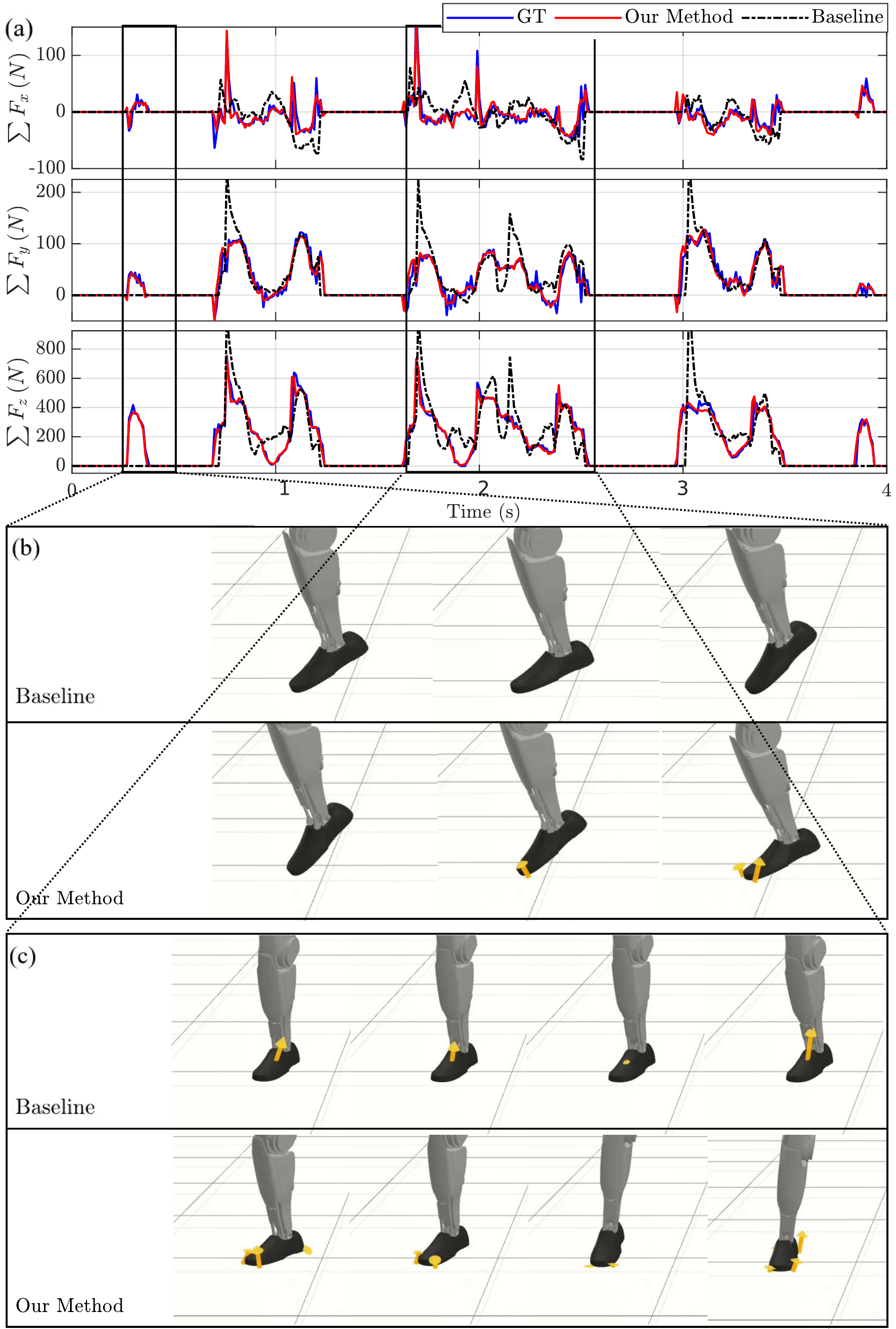}
    \vspace{-10pt}
    \caption{(a) Force-estimation comparison between our method and the contact-constrained estimation baseline for the G1 humanoid (right foot) over a simulated dancing motion. (b) Loss of contact fidelity during partial support at the foot tip. (c) Inaccurate contact reconstructions during rolling and spatially varying foot contact.}
    \label{G1_force_baseline}
    \vspace{-20pt}
\end{figure}

\subsection{Baseline Comparison with Contact-Constrained Estimation}
We compare PRIME with a contact-constrained estimator that optimizes the motion trajectory under a fixed contact sequence, following formulations similar to those in~\cite{PMHE_England,multi_contact_new}. The baseline uses the same full-information estimation objective, measurement model, and cost terms as in~\eqref{eq:FIE}. The key difference lies in the dynamics model and contact constraints. Instead of using the smoothed contact dynamics in~\eqref{eq:contact_smoothed}, the baseline first assigns contact flags by thresholding the end-effector height computed from the measured kinematics. This is consistent with common hardware pipelines that infer end-effector contacts from pressure sensors, estimated external disturbances, or raw kinematic signals, which typically provide only a binary contact indicator. Given the prescribed contact sequence, each active foot contact in the humanoid experiments is modeled by a rigid contact constraint and a spatial friction wrench cone at the mid-foot. The constrained dynamics are written as:
\begin{equation}
    \mathbf{M}(\mathbf{q})\dot{\mathbf{v}}+\mathbf{h}(\mathbf{q},\mathbf{v})
=
\mathbf{B}^{\top}\boldsymbol{\tau}
+
\textstyle \sum_i \mathbf{J}_i(\mathbf{q})^{\intercal}\mathbf{w}_i,
\end{equation}
with the rigid contact acceleration constraint for each active contact $i$:
\begin{equation}
\mathbf{J}_i(\mathbf{q})\dot{\mathbf{v}}
+
\dot{\mathbf{J}}_i(\mathbf{q},\mathbf{v})\mathbf{v}
=
\mathbf{0}.
\end{equation}
Here $\mathbf{J}_i(\mathbf{q})\in\mathbb{R}^{6\times(6+n)}$ is the spatial frame Jacobian of contact i, located at the center of the footpad in the humanoid experiments. The vector $\textstyle \mathbf{w} =
\begin{bmatrix}
f_x &f_y & f_z & \tau_x & \tau_y & \tau_z
\end{bmatrix}^{\intercal}
$ denotes the equivalent spatial wrench applied at the corresponding contact frame. The wrench $\mathbf{w}$ is constrained by the friction wrench cone constraint \cite{contact_wrench_cone}, which can be written as:
\begin{equation}
\begin{array}{ccc}
|f_x| \leq \mu f_z, &
|f_y| \leq \mu f_z, &
f_z > 0, \\[2pt]
|\tau_x| \leq Y f_z, &
|\tau_y| \leq X f_z, &
\tau_{z,\min} \leq \tau_z \leq \tau_{z,\max}.
\end{array}
\end{equation}
\[
\begin{aligned}
\tau_{z,\min}
&=
-\mu(X+Y)f_z
+
|Y f_x-\mu\tau_x|
+
|X f_y-\mu\tau_y|,\\
\tau_{z,\max}
&=
\mu(X+Y)f_z
-
|Y f_x+\mu\tau_x|
-
|X f_y+\mu\tau_y|,
\end{aligned}
\]
where $X$ and $Y$ denote the length and width of the supporting rectangle. Compared with PRIME, this baseline decouples contact detection from estimation by assigning contact modes from kinematic thresholding outside the estimation problem. PRIME instead reconstructs contact status jointly with motion and inertial parameters through its contact dynamics model, avoiding externally prescribed contact flags. Under the fixed-mode formulation, distributed foot contact is further approximated by a single equivalent wrench at the midfoot. Although this approximation is common in control-oriented formulations with prescribed contacts \cite{opt2skill}, it reduces fidelity for estimation, especially during rolling, partial support, and spatially varying foot contact. As shown in Fig.~\ref{G1_force_baseline} and Table~\ref{tab:G1_baseline_RMSE}, the baseline produces substantially worse contact-force reconstruction despite accurate inertial parameters, highlighting the benefit of explicitly modeling contact dynamics.

% Compared with PRIME, this baseline does not explicitly reason about contact status within the estimation problem. Instead, it relies on contact flags obtained from kinematic thresholding and enforces contact through rigid constraints under the prescribed contact mode. With this constraint-based formulation, distributed foot contact is further summarized by a single equivalent wrench at the midfoot. While such an approximation is common for control-oriented formulations with prescribed contact modes, it reduces contact fidelity in estimation, particularly during rolling, partial support, and motions with spatially varying foot contact. These simplifications reduce contact fidelity and lead to substantially worse contact-force reconstruction despite accurate inertial parameters, as shown in Fig.~\ref{G1_force_baseline} and Table. \ref{tab:G1_baseline_RMSE}, highlighting the benefit of explicitly modeling contact dynamics. 

\subsection{Smoothed Contact Dynamics Analysis}
\begin{figure}[t]
  \centering
  \setlength{\abovecaptionskip}{0pt}
  % \includesvg[width=0.90\linewidth]{figure/kappa_force_cost.svg}
  \includegraphics[width=0.90\linewidth]{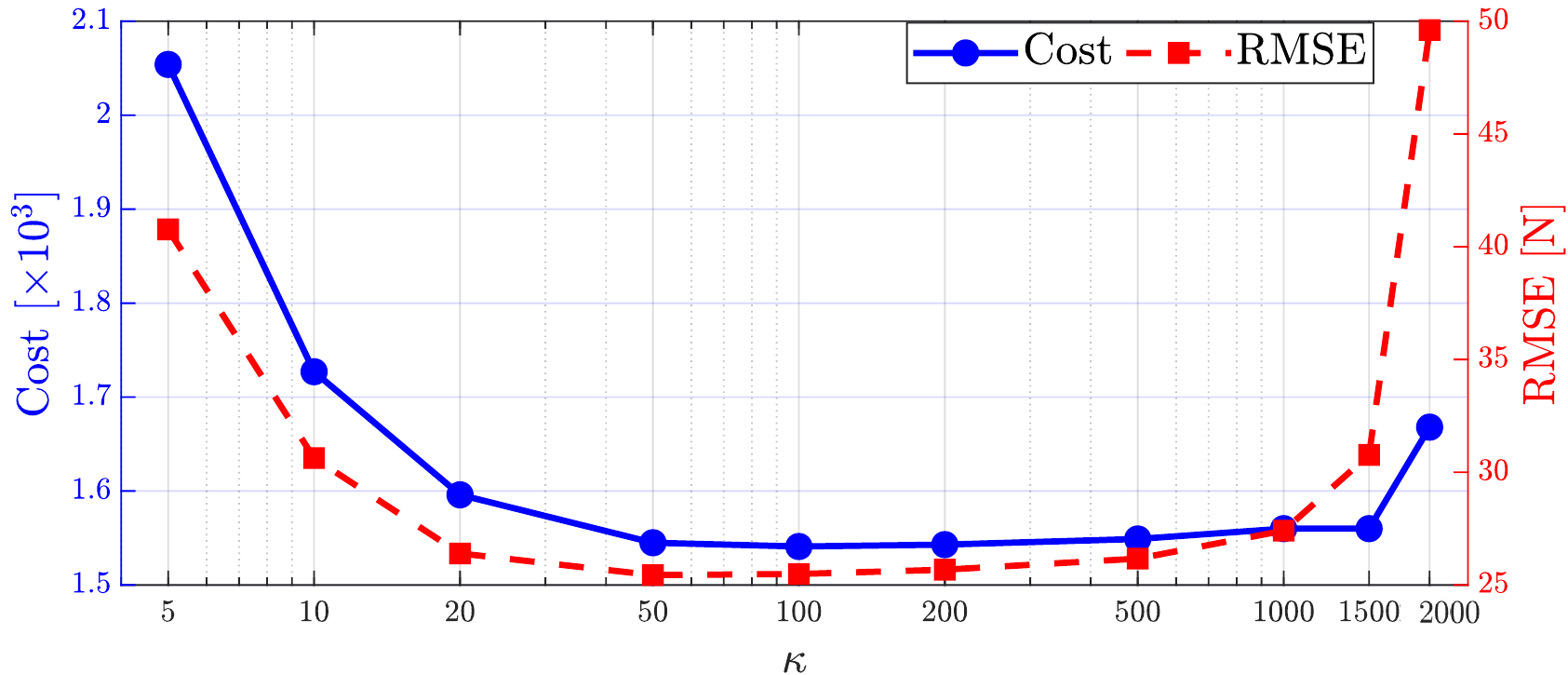}
  \caption{Effects of the smoothing parameter of contact dynamics on the optimization convergence (blue) and the estimation accuracy of contact forces (red) for G1 humanoid in simulation.}
  \label{kappa}
  \vspace{-10pt}
\end{figure}

\begin{figure}[!t]
  \centering
  \setlength{\abovecaptionskip}{0pt}
  \includegraphics[width=0.92\linewidth]{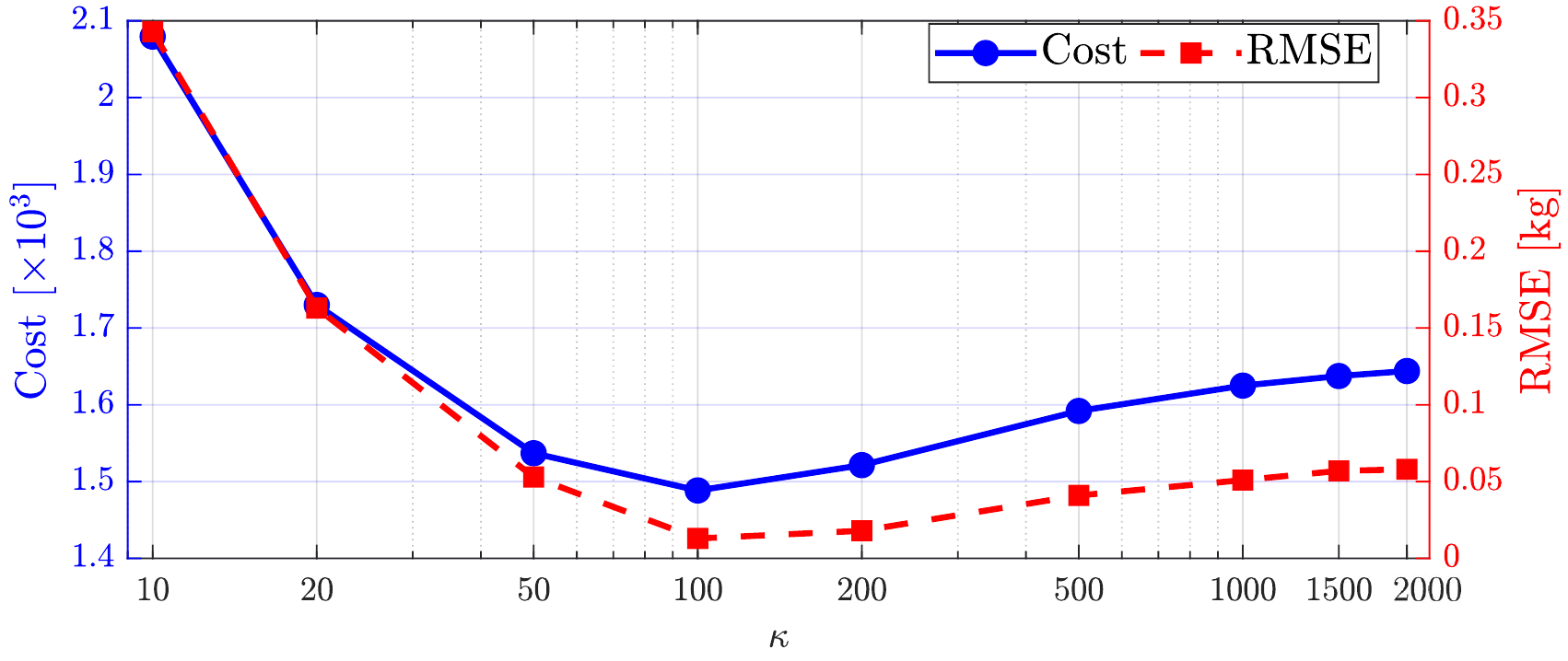}
  % \includesvg[width=0.92\linewidth]{figure/kappa_mass_cost.svg}
  \caption{Effects of the smoothing parameter of contact dynamics on the optimization convergence (blue) and the identification accuracy of torso mass (red) for Go2 quadrupedal in simulation.}
  \label{kappa_id}
  \vspace{-15pt}
\end{figure}

PRIME enables robust reconstruction of latent contact modes by leveraging the smoothed contact dynamics. In particular, the log-barrier approximation of the second-order contact constraints (SOC) as in Section III.C can be interpreted as inducing an implicit ``force field''~\cite{TaoCIMPC}: the barrier introduces a repulsive potential whose influence increases as the solution approaches the constraint boundary, thereby smoothly and differentiably shaping contact engagement and disengagement. As $\kappa \to \infty$, the approximation approaches hard SOC enforcement and recovers rigid, physically consistent contact. For moderate $\kappa$, the relaxation can introduce a force-at-a-distance effect: contact forces may appear smoothly before strict contact activation. This behavior should be interpreted differently in the proposed estimation problem than in its control-side counterpart, contact-implicit control. In control, smoothing artifacts are usually undesirable because the primary goal is task execution with physically feasible trajectories, where contact should be rigidly enforced; the relaxation is therefore often reduced or removed over iterations, mainly to aid numerical convergence. In contrast, PRIME aims to reconstruct motion, contact, and inertial parameters that best explain noisy kinematic measurements and onboard actuator sensing. In this setting, moderate smoothing can improve robustness by reducing sensitivity to contact switching, measurement noise, and data down-sampling. 

We quantify these effects in PRIME by varying $\kappa$ under the same cost settings and evaluating the resulting convergence behavior, force-estimation accuracy, and identification accuracy using simulation data from the G1 and Go2 experiments in Section~V. As shown in Fig. \ref{kappa} and Fig. \ref{kappa_id}, a moderate choice of $\kappa=500$ promotes convergence by providing informative gradients, while maintaining physical consistency with rigid contact dynamics and tolerating noise and contact misalignment introduced by data downsampling. Unless otherwise stated, we use the same $\kappa$ across all experiments. When the hardware measurements are particularly noisy, PRIME can be initialized with a smaller $\kappa$ (stronger smoothing) and then gradually increase $\kappa$ over iterations to the target value, thereby improving convergence while ultimately enforcing physically consistent contact behavior.

\subsection{Computation Profiling}
Although PRIME is currently formulated as an offline long-horizon Full-Information Estimation (FIE) problem for system identification and motion reconstruction, its single-step computation is efficient due to the fully C++ implementation. On a PC equipped with an Intel Core i9-13900 CPU, one quadruped simulation rollout with differentiation takes 80~$\mu$s on average. This is comparable to the 70~$\mu$s differentiation time reported in~\cite{KAISTCIMPC} for a similar contact-rich setting with real-time contact-implicit MPC, although the rollout time is not separately profiled in~\cite{KAISTCIMPC}. Given the similar computational structure of MPC and Moving Horizon Estimation (MHE), this timing suggests that PRIME may be amenable to a real-time receding-horizon formulation. Such an extension would require an appropriate arrival cost to propagate posterior information across sliding windows, especially for the coupled uncertainty among motion states, contact forces, and inertial parameters. We leave this MHE implementation and arrival-cost design for future work, following related frameworks~\cite{KangMHE_RAL}.

\begin{figure}[!htbp]
  \centering
  \setlength{\abovecaptionskip}{0pt}
  % \includesvg[width=0.98\linewidth]{figure/Go2_base.svg}
  \includegraphics[width=0.98\linewidth]{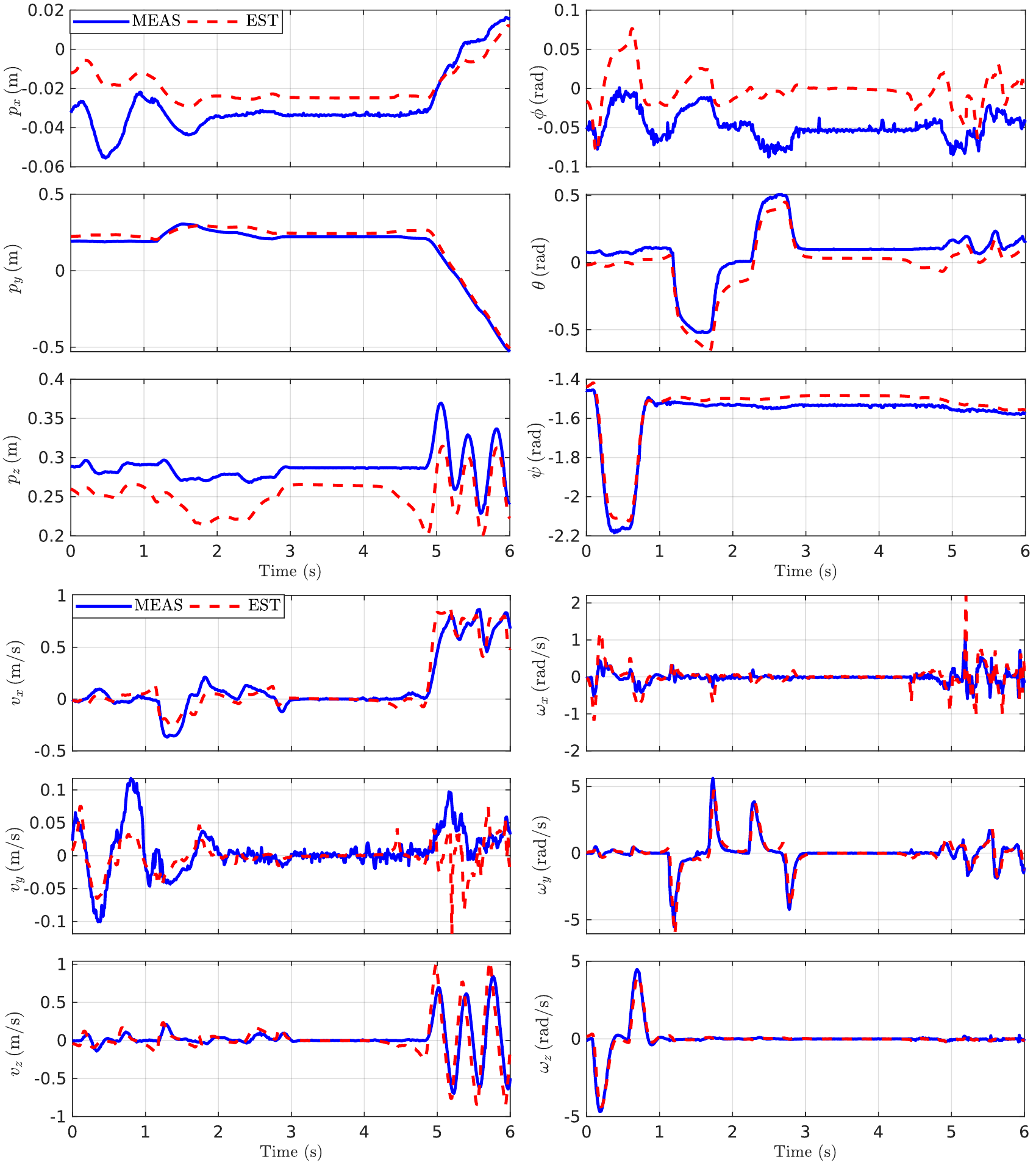}
  \caption{Floating-base estimation of Go2 over diverse motions. The roll component of the mocap measurement orientation is biased due to imperfect calibration.}
  \label{fig:Go2_base}
  \vspace{-5pt}
\end{figure}

\begin{figure}[!htbp]
  \centering
  \setlength{\abovecaptionskip}{0pt}
  % \includesvg[width=\linewidth]{figure/Go2_q.svg}
  \includegraphics[width=1.0\linewidth]{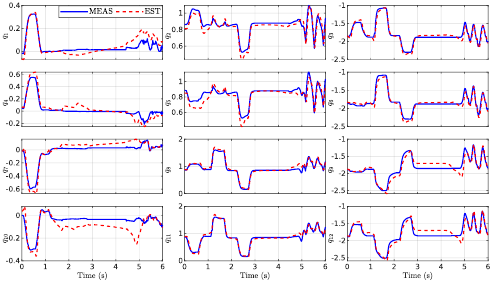}
    \vspace{-10pt}
  \caption{Joint position estimation of Go2 over diverse motion.}
  \label{fig:Go2_q}
  % \vspace{-10pt}
\end{figure}

\begin{figure}[!htbp]
  \centering
  \setlength{\abovecaptionskip}{10pt}
  % \includesvg[width=\linewidth]{figure/Go2_v.svg}
    \includegraphics[width=1.0\linewidth]{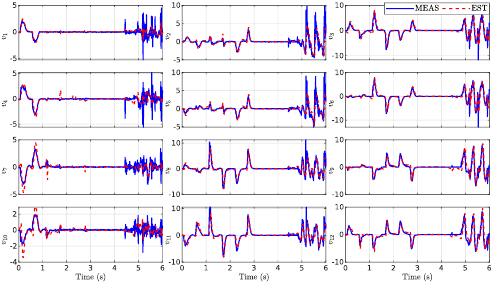}
    \vspace{-20pt}
  \caption{Joint velocity estimation of Go2 over diverse motion.}
  \label{fig:Go2_v}
  \vspace{-10pt}
\end{figure}

\begin{figure}[t]
  \centering
  \setlength{\abovecaptionskip}{-10pt}
  % \includesvg[width=0.93\linewidth]{figure/Go2_tau.svg}
    \includegraphics[width=0.93\linewidth]{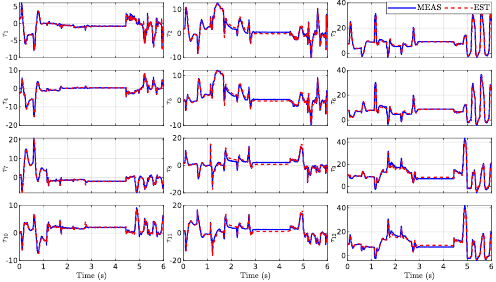}
    \vspace{8pt}
  \caption{Joint torque estimation of Go2 over diverse motion.}
  \label{fig:Go2_tau}
  \vspace{-5pt}
\end{figure}

\begin{figure}[t]
  \centering
  \setlength{\abovecaptionskip}{0pt}
  % \includesvg[width=0.95\linewidth]{figure/g1_base.svg}
  \includegraphics[width=0.95\linewidth]{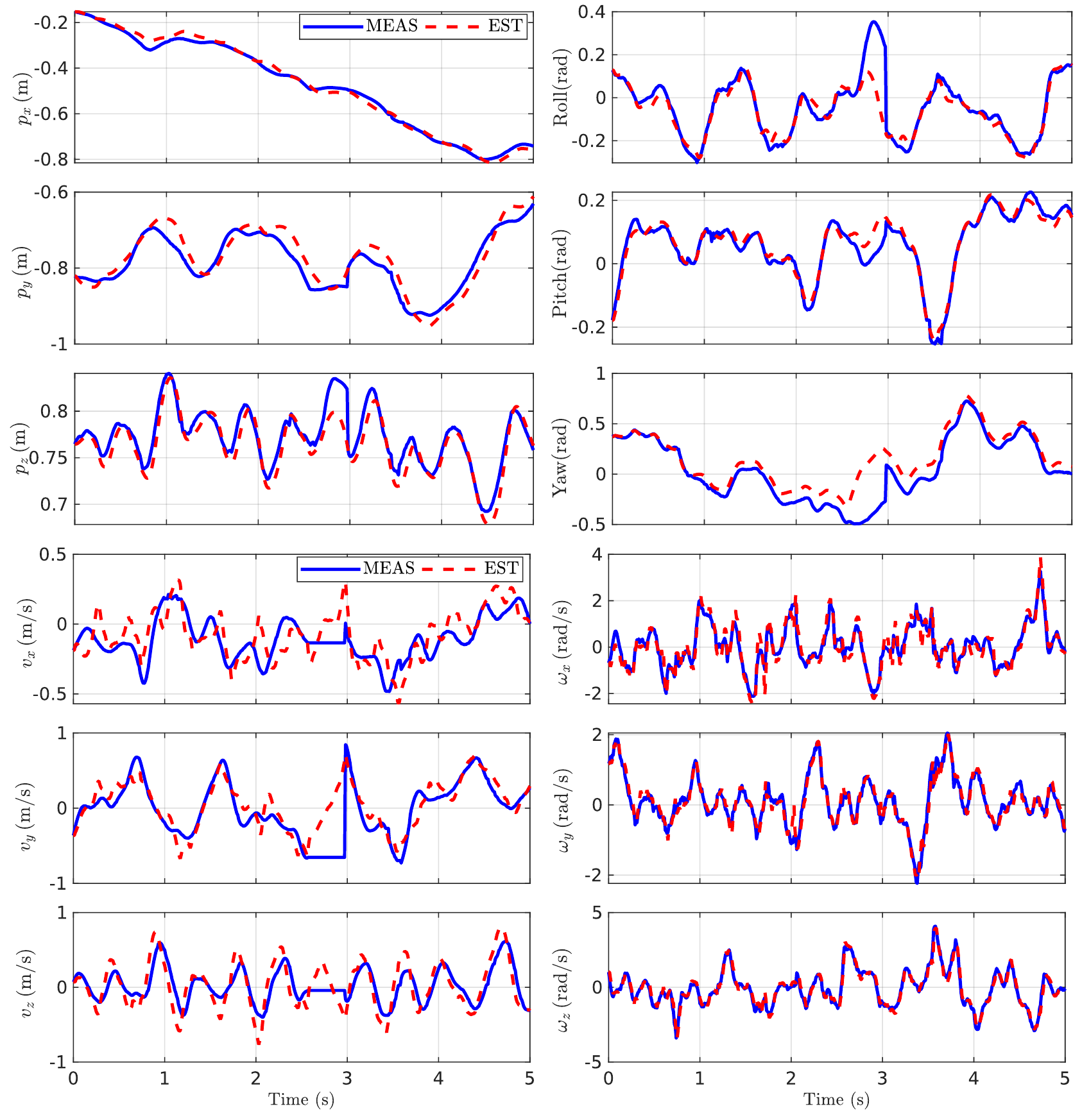}
    \vspace{-5pt}
  \caption{Floating base estimation of G1 humanoid over a dancing motion.}
  \label{fig:G1_base}
  \vspace{-4pt}
\end{figure}

\begin{figure}[t]
  \centering
  \setlength{\abovecaptionskip}{0pt}
  % \includesvg[width=0.93\linewidth]{figure/G1_q.svg}
    \includegraphics[width=0.93\linewidth]{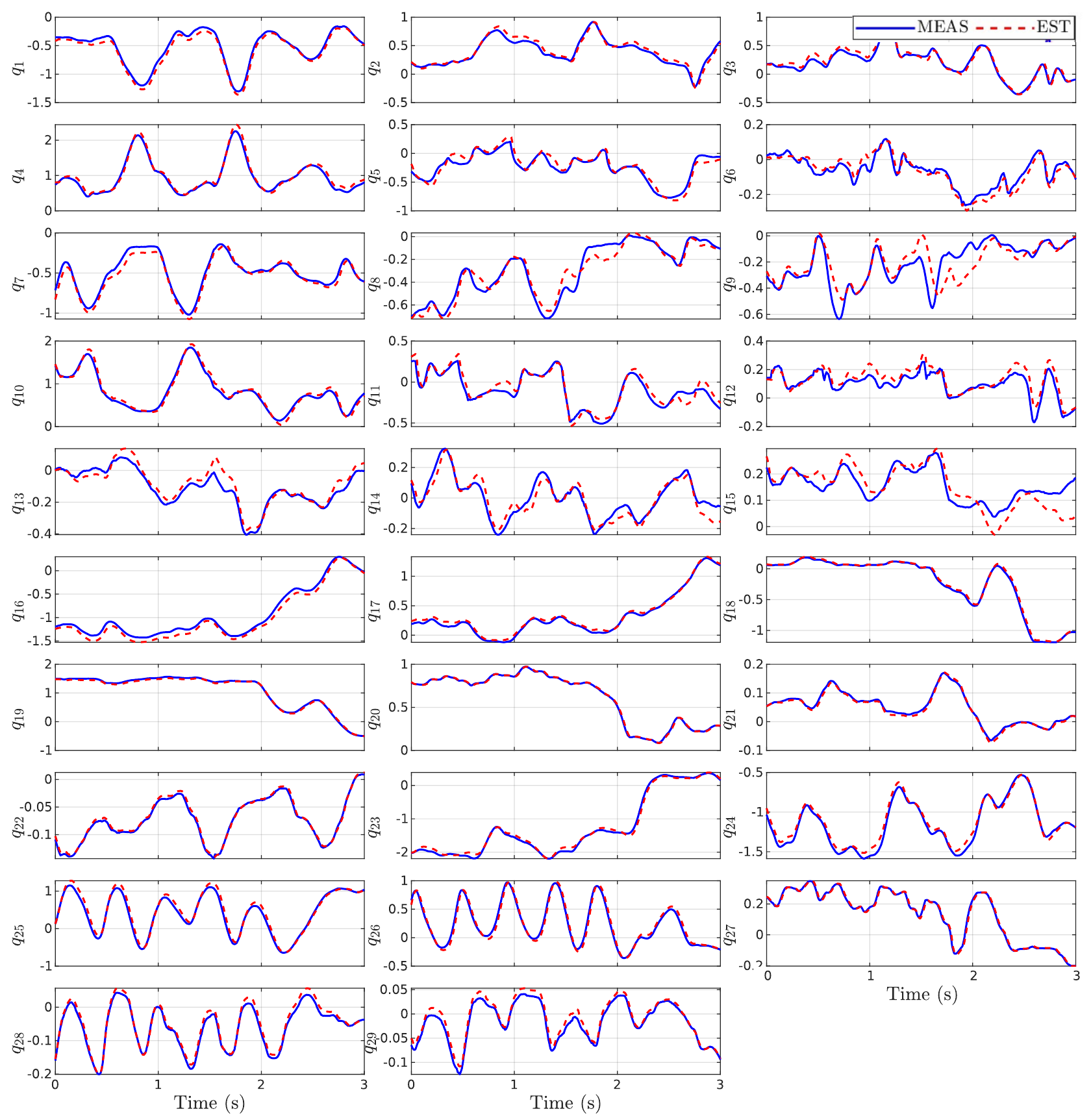}

  \caption{Joint position estimation of G1 humanoid over a dancing motion.}
  \label{fig:G1_p}
  % \vspace{-10pt}
\end{figure}

\clearpage

\begin{figure}[H]
  \centering
  \setlength{\abovecaptionskip}{0pt}
  % \includesvg[width=\linewidth]{figure/G1_v.svg}
    \includegraphics[width=1.0\linewidth]{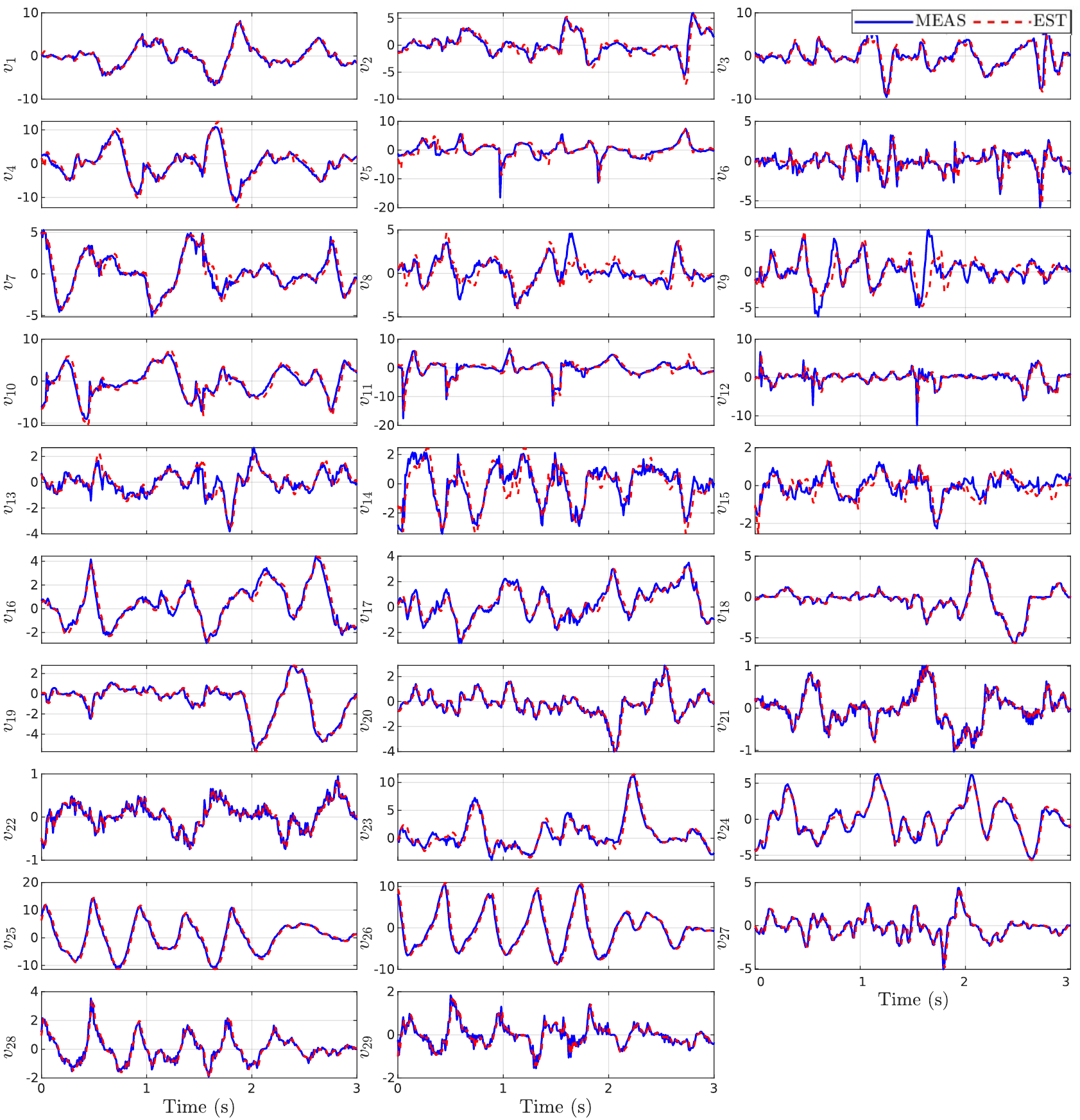}
  \caption{Joint velocity estimation of G1 humanoid over a dancing motion.}
  \label{fig:G1_v}
  \vspace{-8pt}
\end{figure}

\newpage

\begin{figure}[H]
  \centering
  \setlength{\abovecaptionskip}{0pt}
  % \includesvg[width=\linewidth]{figure/G1_tau.svg}
      \includegraphics[width=1.0\linewidth]{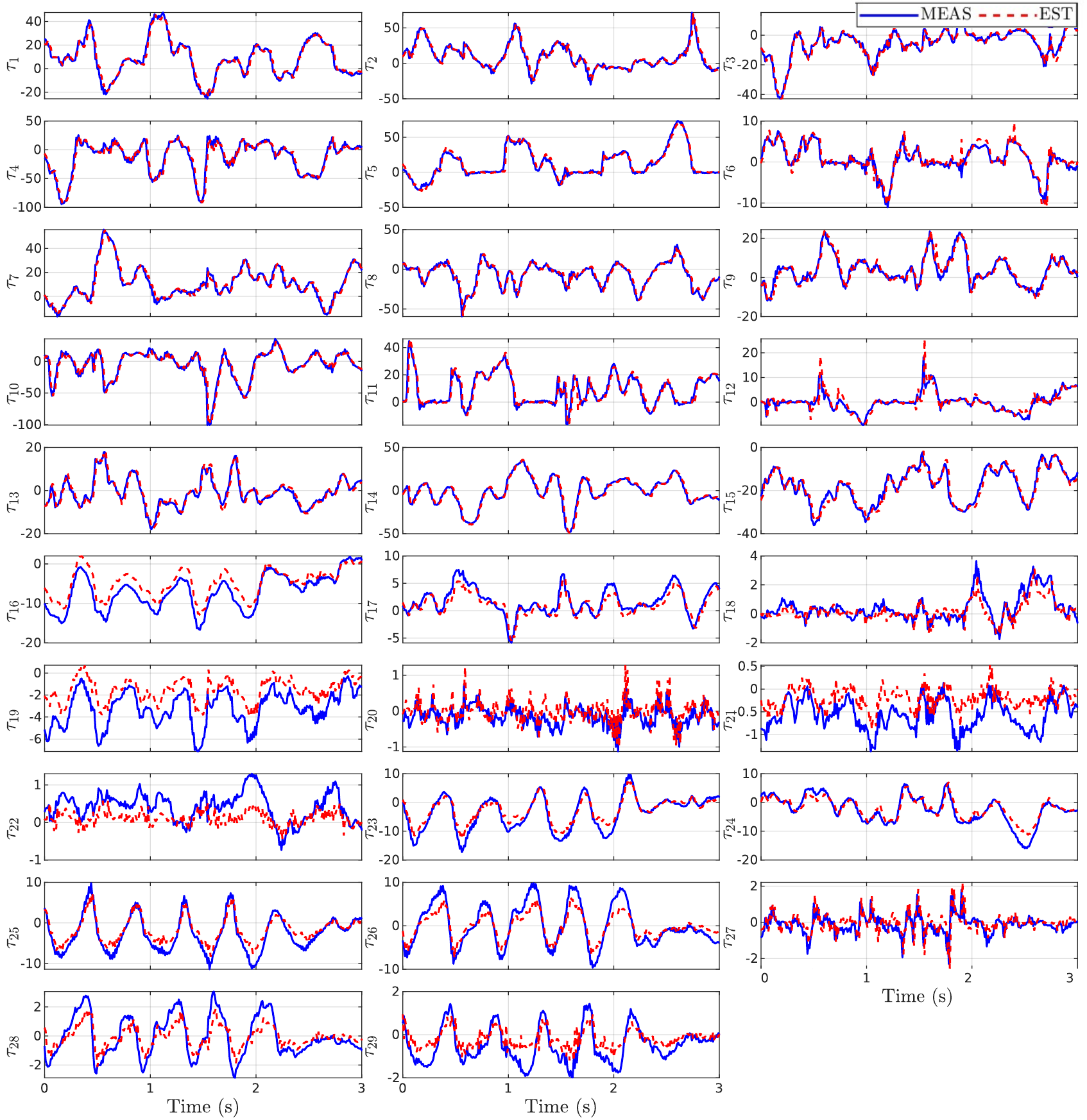}
  \caption{Joint torque estimation of G1 humanoid over a dancing motion.}
  \label{fig:G1_tau}
  \vspace{-8pt}
\end{figure}

\end{document}